\documentclass[runningheads]{llncs}

\usepackage{eccv}

\usepackage{eccvabbrv}

\usepackage{booktabs}

\usepackage{graphicx}
\usepackage{multicol,lipsum}
\usepackage{amsmath}
\usepackage{amssymb}
\usepackage{listings}
\usepackage[export]{adjustbox}
\usepackage{epsfig}
\usepackage{algorithm}
\usepackage{algpseudocode}
\usepackage{enumitem}
\usepackage{color, colortbl}
\usepackage{multirow}
\usepackage{url}
\usepackage{arydshln}
\usepackage{soul}
\usepackage{capt-of}
\usepackage{xstring}
\usepackage{wrapfig}
\usepackage{color}
\usepackage{cuted}
\usepackage{fontawesome5}

\newcommand\blfootnote[1]{%
  \begingroup
  \renewcommand\thefootnote{}\footnote{#1}%
  \addtocounter{footnote}{-1}%
  \endgroup
}
\newcommand*{\img}[1]{%
    \raisebox{-.15\baselineskip}{%
        \includegraphics[
        height=\baselineskip,
        width=\baselineskip,
        keepaspectratio,
        ]{#1}%
    }%
}
\definecolor{codegreen}{rgb}{0,0.6,0}
\definecolor{codegray}{rgb}{0.5,0.5,0.5}
\definecolor{codepurple}{rgb}{0.58,0,0.82}
\definecolor{backcolour}{rgb}{0.95,0.95,0.92}
\lstdefinestyle{mystyle}{
  backgroundcolor=\color{backcolour}, commentstyle=\color{codegreen},
  keywordstyle=\color{magenta},
  numberstyle=\tiny\color{codegray},
  stringstyle=\color{codepurple},
  basicstyle=\ttfamily\footnotesize,
  breakatwhitespace=false,         
  breaklines=true,                 
  captionpos=b,                    
  keepspaces=true,                 
  numbers=left,                    
  numbersep=5pt,                  
  showspaces=false,                
  showstringspaces=false,
  showtabs=false,                  
  tabsize=2,
}
\newcommand{\modelname}{$\mathbb{MST}_\mathbb{MIXER}$}
\definecolor{myblue}{RGB}{0,0,128}
\definecolor{mypurple}{RGB}{219,175,255}
\definecolor{mypurpletext}{RGB}{171,69,255}
\definecolor{mygreen}{RGB}{121,217,154}
\definecolor{mygold}{RGB}{174,126,21}

\definecolor{myblue}{RGB}{0,0,128}
\definecolor{myorange}{RGB}{255,227,200}
\definecolor{mygray}{RGB}{231,231,231}
\definecolor{blue_ref}{RGB}{0,0,128}
\DeclareMathOperator*{\argmax}{arg\,max}

\newcommand{\rom}[1]{\uppercase\expandafter{\romannumeral #1\relax}}
\usepackage{pifont}
\newcommand{\cmark}{\ding{51}}%
\newcommand{\xmark}{\ding{55}}%

\usepackage[accsupp]{axessibility}  

\usepackage[pagebackref,breaklinks,colorlinks,citecolor=eccvblue]{hyperref}

\usepackage{orcidlink}

\begin{document}

\title{Multi-Modal Video Dialog State Tracking in the Wild}

\titlerunning{\modelname \img{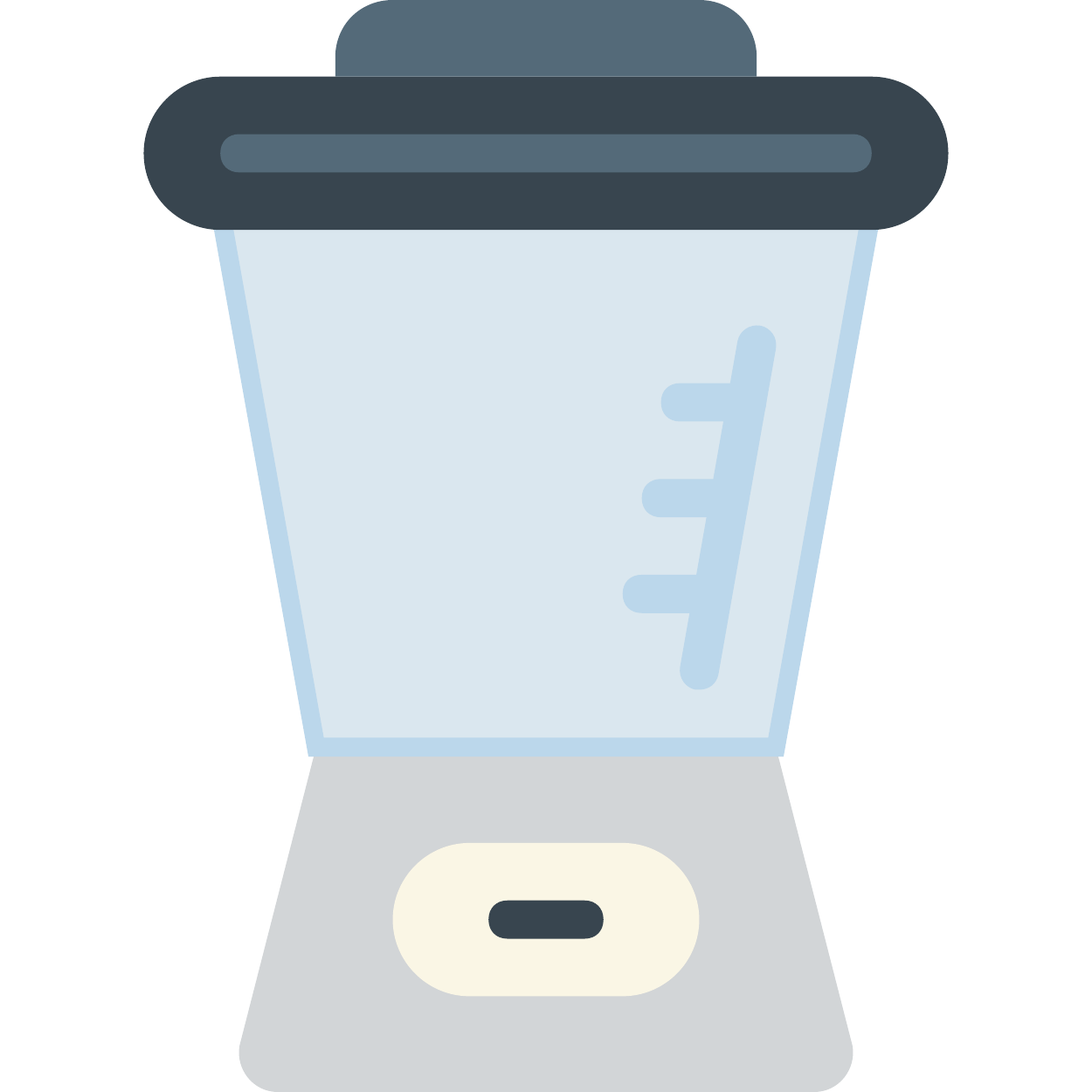}}

\author{Adnen Abdessaied \orcidlink{0000-0002-9489-6340} \and
Lei Shi\orcidlink{0000-0003-1628-1559} \and
Andreas Bulling \orcidlink{0000-0001-6317-7303}}

\authorrunning{A.~Abdessaied et al.}

\institute{University of Stuttgart, Germany\\
\email{\{adnen.abdessaied, lei.shi, andreas.bulling\}@vis.uni-stuttgart.de}\\
\img{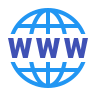} \url{https://perceptualui.org/publications/abdessaied24_eccv}
}

\maketitle

\begin{abstract}
  We present \modelname \img{figures/mixer_icon.pdf} -- a novel video dialog model operating over a generic multi-modal state tracking scheme.
Current models that claim to perform multi-modal state tracking fall short in two major aspects:
(1) They either track only one modality (mostly the visual input)
or (2) they target synthetic datasets that do not reflect the complexity of real-world in-the-wild scenarios.
Our model addresses these two limitations in an attempt to close this crucial research gap.
Specifically, \modelname\, first tracks the most important constituents of each input modality.
Then, it predicts the \textit{missing} underlying structure of the selected constituents of each modality by learning local latent graphs using a novel multi-modal graph structure learning method.
Subsequently, the learned local graphs and features are parsed together to form a global graph operating on the mix of all modalities, further refining its structure and node embeddings.
Finally, the fine-grained graph node features are used to enhance the hidden states of the backbone Vision-Language Model (VLM).
\modelname\, achieves new state-of-the-art results on \textit{five} challenging benchmarks.

\end{abstract}

\begin{figure}[!h]
    \begin{minipage}{1\linewidth}
        \centering
        \scalebox{0.55}[0.55]{
            \includegraphics[width=\textwidth]{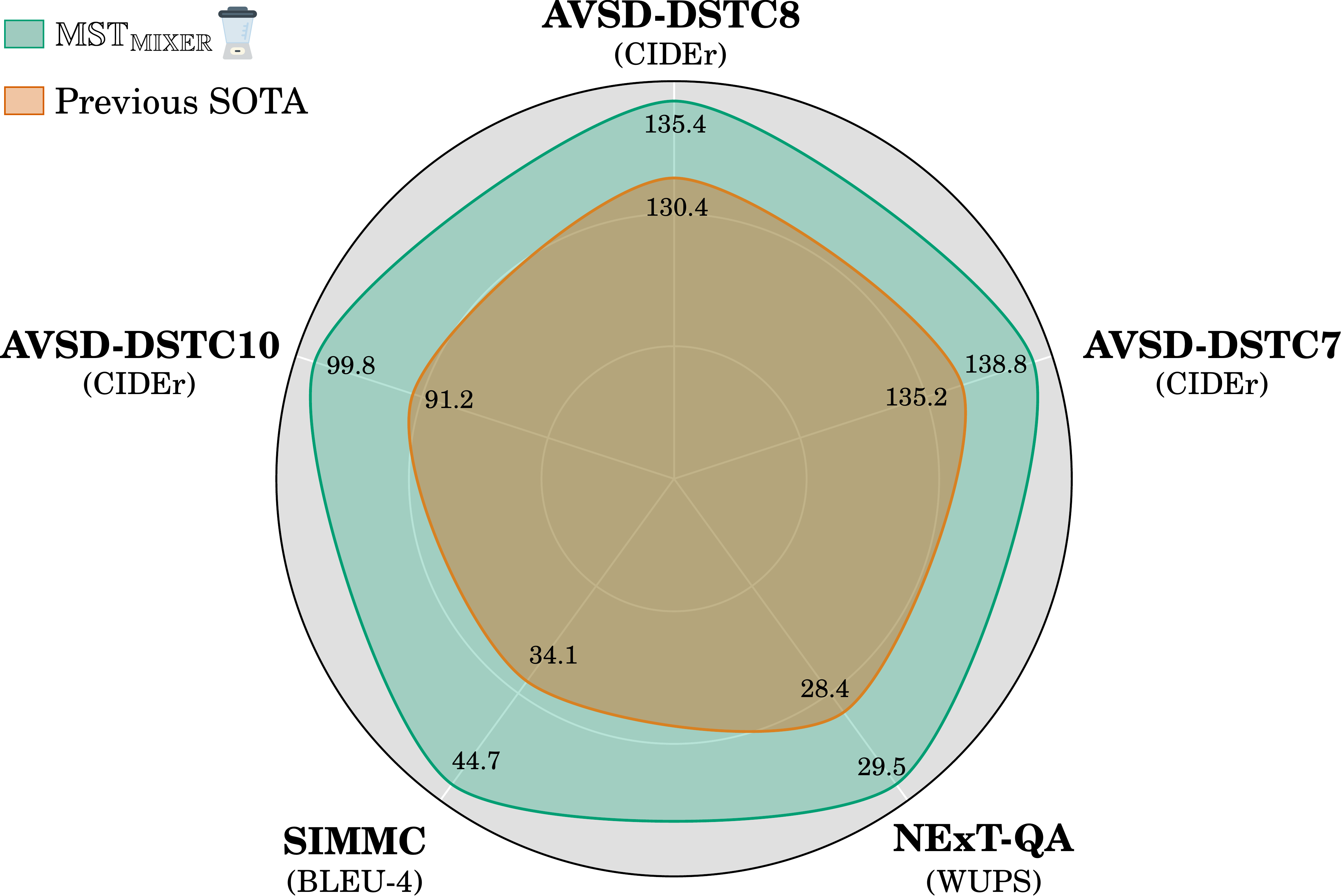}
        }
        \caption{\modelname \img{figures/mixer_icon.pdf} achieves SOTA results on various video-language tasks.
        }
        \label{fig:teaser}
    \end{minipage}
\end{figure}

\section{Introduction}
Multi-modal tasks at the intersection of computer vision and natural language  
processing have been introduced to develop intelligent agents capable of assisting humans in understanding a visual premise through language. 
Among these tasks, video dialog is considered to be one of the most challenging.
In contrast to visual~\cite{vqa} and video~\cite{xu2017video} question answering, which only require reasoning about a single question, video dialog models have to reason over the entire dialog history in addition to the current question.
Furthermore, in contrast to visual dialog~\cite{visdial}, video dialog involves reasoning over a video instead of a static image.
Thus, a crucial part of a video dialog model is Dialog State Tracking (DST), which was originally introduced to track and update users' goals in the form of dialog states \cite{lee-etal-2019-sumbt,wu-etal-2019-transferable}.
Nowadays, it is broadly used when a model keeps track of what it believes to be relevant for answering the question at hand.

Until now, research on DST has been predominately uni-modal in the form of slot-filling tasks \cite{mrksic-etal-2017-neural,xu-hu-2018-end,Le2020Non-Autoregressive} where the slots and slot values are constrained by a knowledge domain (e.g. hotel domain) and database schema (e.g. tabular data).
However, the current landscape of the field necessitates extending to a multi-modal framework.
Current models that claim to perform multi-modal state tracking fall short in two major aspects:
(1) Some works track the constituents of only one modality to help the model focus on the most salient ones within a multi-model context (e.g. video dialog \cite{mou2020multimodal}, visual dialog \cite{pang2020visual}, image retrieval \cite{guo2018dialog}, recommender systems \cite{rec_sys}) rendering their state tracking approach uni-modal.
More recently, Le \etal \cite{Le2022} have proposed VDTN, which extended the slot-filling paradigm to predict the visual attributes of CATER objects \cite{cater} from a pool of pre-defined textual values, but their approach suffers from the same aforementioned limitation.
(2) Other works \cite{simmc,simmc0,abdessaied_olvit} have moved closer to performing multi-modal state tracking but have been limited to synthetic datasets that do not reflect the complexity of real-world scenarios.

We present \modelname\, as a step towards addressing the aforementioned limitations.
Specifically, \modelname\, uses a backbone VLM and attention-based modality-specific tracking blocks to identify the most relevant constituents of each modality.
Then, it uses a multi-modal GNN-based approach to learn the missing underlying structure between the mix of modalities in the form of latent graphs.
Finally, it uses the fine-grained GNN features to enhance the hidden states of the backbone VLM to answer the question at hand more efficiently.
To summarize, the contributions of our work are three-fold:
(1) We propose \modelname -- a novel video dialog model that, unlike previous works, performs multi-modal state tracking on each input modality separately.
Our model is generic by nature and could be easily adapted to deal with
a wide range of tasks and datasets.  
(2) We equip our model with a novel divide-and-conquer GNN-based mechanism
that dynamically learns the missing underlying structure of the mix of all modalities. 
First, it selects the most important constituents of each modality and learns their respective local structures using latent graphs.
Then, it parses all individual graphs and features into a global modality-agnostic graph to further refine its structure and node features that we use to enhance the hidden states of the backbone VLM. 
(3) As seen in \autoref{fig:teaser}, \modelname\, sets new state-of-the-art results across a broad range of video-language tasks.

\section{Related Work}
\subsubsection{Video Dialog.}
Video dialog has emerged as a natural extension to visual question answering \cite{vqa}, video question answering \cite{videoqa}, and visual dialog \cite{visdial}.
Almari \etal \cite{avsd} proposed AVSD -- one of the first video dialog datasets
based on the Charades videos \cite{charades}, which has become the default dataset for the task.
Later works \cite{le-hoi-2020-video,9376902} achieved new state-of-the-art results by leveraging pre-trained large language models \cite{Lewis2020,radford2019language} and fine-tuning them on the downstream video dialog task.
Others used GNNs to perform reasoning on the dialog history \cite{le2021learning} or on the visual scene \cite{scga} in an attempt to improve performance.
Pham \etal \cite{pham2022video} proposed an object-centric model to track object-associated dialog states upon receiving new questions.
Inspired by the success of neural module networks \cite{nmn,nmn_qa}, Le \etal \cite{vgnmn} introduced VGNMN to model the information retrieval process in video-grounded language tasks as a pipeline of neural modules.
More recently, Yoon \etal \cite{tham} introduced a text hallucination mitigation framework based on a hallucination regularization loss.

Despite the high multi-modality of the task in general and the AVSD dataset in particular, all previous works missed out on the idea of performing explicit multi-modal dialog state tracking.
Instead, they focused on general vanilla attention methods that particularly tracked \textit{only} one modality (mostly the visual input) at the expense of the others.
\modelname\, closes this gap by performing multi-modal state tracking on each input modality separately.

\subsubsection{Dialog State Tracking.}
Traditional state tracking approaches predicted slot values (e.g. meals offered by a restaurant) from a pre-defined set at each dialog, which is conditioned on some context.
As a result, these approaches remained predominately uni-modal
even though they were applied within a multi-modal context (e.g., video dialog \cite{mou2020multimodal}, visual dialog \cite{pang2020visual}, image retrieval \cite{guo2018dialog}, recommender systems \cite{rec_sys}).
However, the current landscape of dialog research necessitates the transition to multi-modal dialog state tracking to cope with the complexity of recent datasets.
Some works have already been proposed to address this problem.
For example, SIMMC \cite{simmc0, simmc} was introduced to develop agents capable of helping a human in a shopping scenario and, therefore, need to track the multi-modal state of the dialog to fulfill its task efficiently.
More recently, Le \etal \cite{Le2022} suggested performing video dialog state tracking by extending the slot-filling task to predict predefined attributes of CATER \cite{cater} objects, limiting their approach to only the DVD dataset \cite{dvd}. 

As such, all of these works focused only on synthetic and automatically generated datasets.
To the best of our knowledge, \modelname\, is the first model to perform genuine
multi-modal state tracking in the wild for video dialog by being able to deal with complex real-world scenarios.

\subsubsection{Graph Structure Learning.}
Early works on graph structure learning leveraged bilevel programming \cite{colson2007overview} to simultaneously learn GNN parameters and topology \cite{franceschi2019learning}.
Yu \etal \cite{yu2019dag} proposed applying the linear structure equation model in conjunction with a variational autoencoder \cite{vae} to learn directed acyclic graphs.
Subsequently, Elinas \etal \cite{elinas2020variational} suggested using a stochastic variational inference model to jointly estimate the graph posterior and the GNN parameters.
Chen \etal  \cite{chen2020iterative} proposed iteratively refining the graph topology in an end-to-end manner using graph similarity metric learning.
Wu \etal \cite{wu2022nodeformer} suggested an all-pair message passing method to propagate signals between arbitrary nodes for classification efficiently.

Our method differs from the aforementioned works in three distinct aspects:
(1) We propose a novel multi-modal graph structure learning method that relies on a two-stage divide-and-conquer procedure that first predicts local modality-specific latent graphs before tackling the global graph consisting of the mix of all available modalities. 
(2) We use our graph learning approach to enhance the hidden states of a backbone VLM.
(3) Instead of dealing with uni-modal graph-based tasks (node, edge, or graph classification), we investigate the effect of our method on the multi-modal, non-graph-related downstream task of video dialog.

\begin{figure}[!t]
    \begin{minipage}{1\linewidth}
        \centering
        \scalebox{0.97}[0.97]{
            \includegraphics[width=\textwidth]{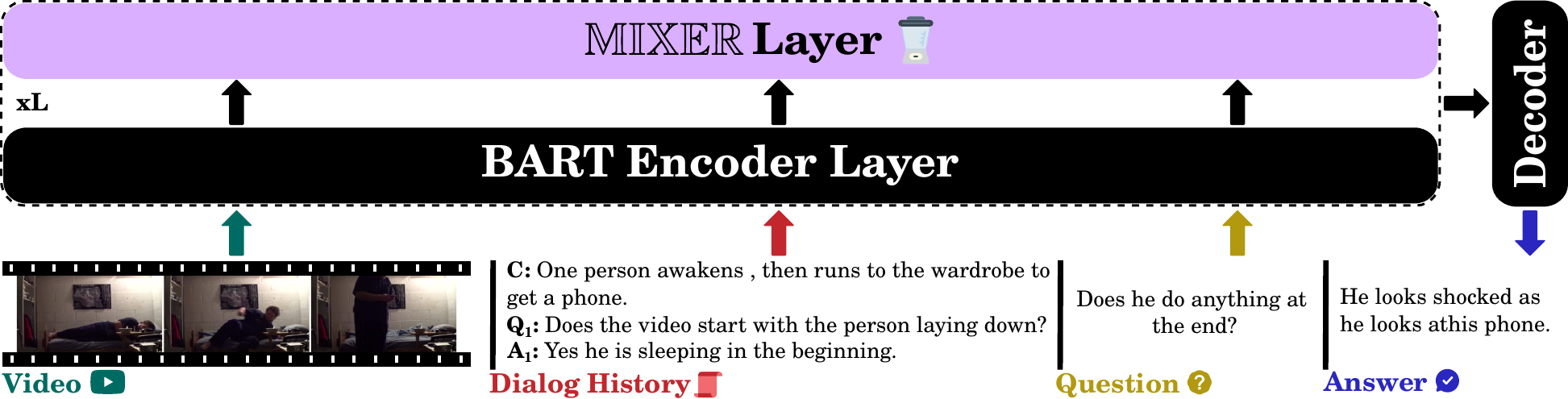}
        }
        \caption{
        \modelname \img{figures/mixer_icon.pdf} 
        takes a video \img{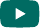}, a dialog history \img{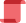}, and a question \img{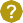} as input and autoregressively generates an answer \img{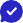} as output.
        It uses a BART backbone adapted to deal with multi-modal input features and enhanced via our graph-based mixing approach. 
        }
        \label{fig:overview}
    \end{minipage}
\end{figure}
\section{Method}

\subsection{Problem Formulation}
Given a question $\texttt{Q}_\texttt{t}$ grounded on a video $\texttt{V}$ at $\texttt{t}$-th dialog turn, a dialog history  $\texttt{H}_\texttt{t} = \{\texttt{C}, \texttt{(Q}_\texttt{1}, \texttt{A}_\texttt{1}\texttt{)}, ..., \texttt{(Q}_{\texttt{t-1}}, \texttt{A}_{\texttt{t-1}} \texttt{)}\}$ composed of previous question-answer pairs and a video caption $\texttt{C}$, a video dialog model is tasked of autoregressively generating a free-form answer $\texttt{A}_\texttt{t}$ to the question at hand, i.e. each answer token $\texttt{a}_\texttt{t}^\texttt{i}$ satisfies    
\begin{equation}
    \texttt{a}_\texttt{t}^\texttt{i} = \displaystyle \argmax_{\texttt{a} \in \mathcal{V}} \left[ P\left(\texttt{a} | \texttt{V}, \texttt{Q}_\texttt{t}, \texttt{H}_\texttt{t},  \texttt{A}_\texttt{t}^{<\texttt{i}} \right)\right],%
\end{equation}
where $\texttt{A}_\texttt{t}^{<\texttt{i}}$ and $\mathcal{V}$ denote the previously predicted answer tokens and the vocabulary, respectively.

\subsection{Input Representation Learning}
As can be seen from \autoref{fig:overview}, \modelname\, is based on BART \cite{Lewis2020} and adapted to handle data from multiple input modalities.

\subsubsection{Visual Representations.}
As it is standard for this task, the visual representations are extracted for a given video using I3D-rgb and I3D-flow models \cite{carreira2017quo} pre-trained on YouTube videos and the Kinetics dataset \cite{kay2017kinetics}.
Formally, a video \texttt{V} is first split into $l_\textrm{v}$ segments using a sliding window of $n$ frames.
Then, each segment $S = \{f_1, f_2, ..., f_n\}$, where $f_i$ represents one video frame, are fed to the pre-trained I3D models to extract the $d_v$-dimensional video features $V_\textrm{rgb}, V_\textrm{flow} \in \mathbb{R}^{l_\textrm{v} \times d_\mathrm{v}}$.
Finally, we extracted object features $V_\textrm{sam}\in \mathbb{R}^{l_\textrm{v} \times d_\mathrm{s}}$ from the middle frame of the video using SAM \cite{sam}. 
We mapped these features to match the hidden dimension $d$ of BART using linear projections with weights matrices $W_\textrm{rgb}, W_\textrm{flow}, W_\textrm{sam}$.

\subsubsection{Audio Representations.}
Similar the previous works \cite{tham,le2021learning,9376902}, we used audio features extracted from a pre-trained VGGish model \cite{simonyan2015a}.
Since video and audio are synchronous, the same splits were used to generate the $d_a$-dimensional audio features $A_\textrm{vggish} \in \mathbb{R}^{l_\textrm{v} \times d_a}$.
As for the video feature, we mapped the audio features to the BART embedding space using a linear projection with a weight matrix $W_{a} \in \mathbb{R}^{d\times d_a}$.
We refer to \cite{avsd_baseline} for further details about feature extraction.

\subsubsection{Textual Representations.}
We used the dialog history composed of the video caption, the previous question-answer pairs, and the current question as additional input to the encoder.
We separated each segment with the special token \texttt{<\texttt{/s}>}.
Subsequently, we embedded their concatenation into a dense representation $T = [T_\textrm{H}, T_\textrm{Q}] \in \mathbb{R}^{l_\mathrm{txt} \times d}$ using a word embedding matrix $W_\mathrm{txt}\in \mathbb{R}^{|\mathcal{V}| \times d}$, where $l_\textrm{txt}$, $\mathcal{V}$, $T_\textrm{H}$, and $T_\textrm{Q}$ are the length of the textual input, the vocabulary, the dense representation of the history and question, respectively.
Finally, we input a shifted ground truth into the decoder and embed it using the same word matrix.

\subsubsection{State Tokens.}
We inserted special \textit{state} tokens $\texttt{<s}_{\texttt{i}}\texttt{>}$ at the beginning of each modality ($V_\textrm{rgb}, V_\textrm{flow}, V_\textrm{sam}, A_\textrm{vggish}, T_\textrm{H}, T_\textrm{Q}$) and used them to keep track of the most relevant constituents. 

\begin{figure}[!t]
    \begin{minipage}{1\linewidth}
        \centering
        \scalebox{0.98}[0.98]{
            \includegraphics[width=\textwidth]{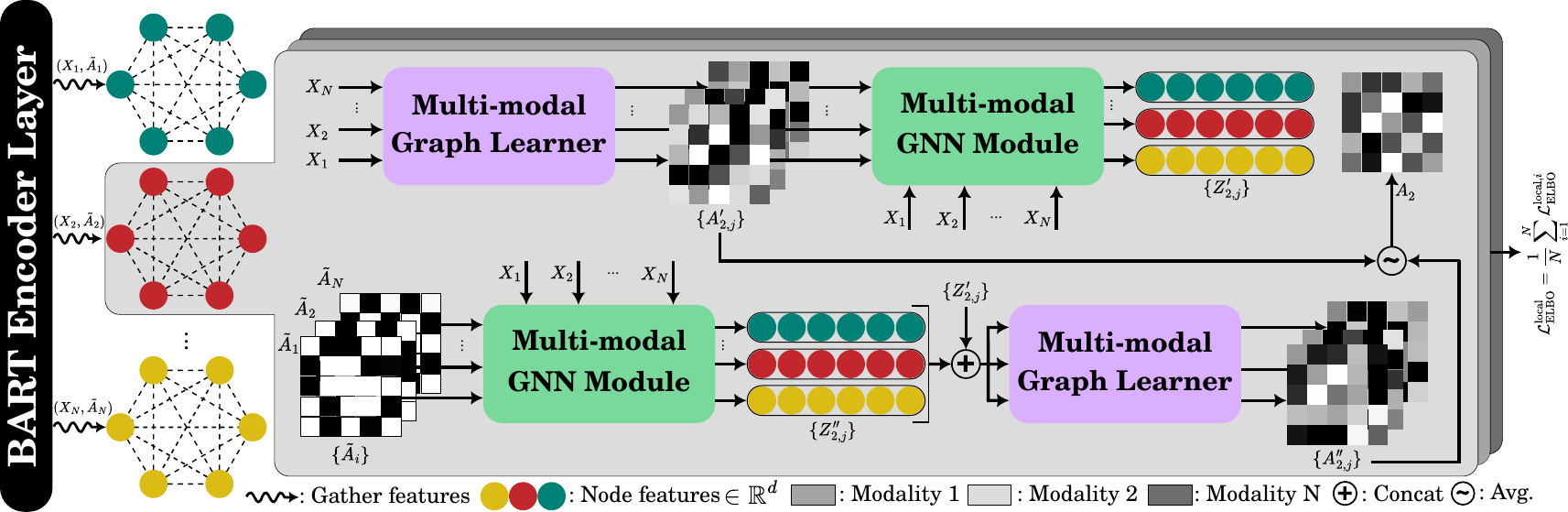}
        }
        \caption{
            In Stage \rom{1}, \modelname\, first gathers multi-modal features $\{X_i\}$ from the previous BART layer and computes their respective initial local structures $\{\tilde{A}_I \}$.
            Then, it simultaneously learns the local latent multi-modal graphs and refines the features using a two-stream framework, i.e., $\{A'_{i,j}, A''_{i,j}\}_j$ and $\{Z'_{i,j}, Z''_{i,j}\}_j$, respectively.
            Finally, it outputs the final multi-modal latent graph $A_i$ used to compute the local ELBO loss  $\mathcal{L}_\textrm{ELBO}^\textrm{local} = \frac{1}{N}\sum_{i=1}^N  \mathcal{L}_\textrm{ELBO}^{\textrm{local}, i}$.
        }
        \label{fig:method_stage_1}
    \end{minipage}
\end{figure}
\subsection{\modelname: Multi-Modal Feature Mixing}
The main idea of \modelname\, is to keep track of the most relevant constituents at different semantic levels (e.g. across modalities and encoder layers) and use them to refine the multi-modal state of the model.
Specifically, we insert a $\mathbb{MIXER}$ layer after every $\Delta$ encoder layer. 
Our approach follows a two-stage divide and conquer scheme where we first learn the underlying \textit{local} structures of the individual modalities before learning the \textit{global} inter-modal structure of the mix of all available modalities.
We posit that directly learning the latter might be daunting for such a high multi-modal task.

\subsubsection{Multi-Modal Feature Tracking.}
We take advantage of the special state tokens $\texttt{<s}_{\texttt{i}}\texttt{>}$ to keep track of the most relevant modality-specific features at different embedding levels of the encoder.
Specifically, for each modality, we select the $K$ tokens with the highest attention values concerning the respective state token, i.e. 
\begin{equation}
    X_i = \mathrm{top}_K(\alpha_{\textrm{avg}}( h_{\texttt{<s}_{\texttt{i}}\texttt{>}}, H_i )) \in \mathbb{R}^{K\times d},
\end{equation}
where $\alpha_{\textrm{avg}}( h_{\texttt{<s}_{\texttt{i}}\texttt{>}}, H_i )$ is the attention values between the state embedding and the remaining tokens embeddings $H_i$ of the $i-$th modality averaged across heads.

\subsubsection{Mixing Stage \rom{1} (Divide).}
We posit that the selected features $\{X_i\}$ of each modality encapsulate rich information that could be leveraged to improve the learning capabilities of our model.
A viable approach is to take advantage of the power of GNNs to refine these features based on their local structures, as prior works have highlighted the merit of integrating GNNs with transformer-based models \cite{Abdessaied_2024_WACV,Yang2021,ying2021do}.
However, the underlying structures that govern $\{X_i\}$ are missing in our case.
To this end, we propose a novel multi-modal graph structure learning approach that simultaneously learns the graph weights and the adjacency matrix in the form of latent graphs.
We posit that we can split the adjacency matrix $A_i$ of the $i-$th modality into an initial (observable) part $\tilde{A}_i$ and a missing (sought-after) part $A_i'$ where $\tilde{A}_i$ is a \textit{binary matrix} constructed using a $k$NN ($k=4$) approach based on $X_i$.
Thus,

\begin{align}
    P(X_i, A_i) &= P(A_i|X_i) P(X_i) \\
    & = P(A_i', \tilde{A}_i|X_i) P(X_i).
\end{align}

\begin{figure}[tb]
  \centering
  \begin{subfigure}{0.62\linewidth}
    \centering

    \scalebox{0.98}[0.98]{
        \includegraphics[width=\textwidth]{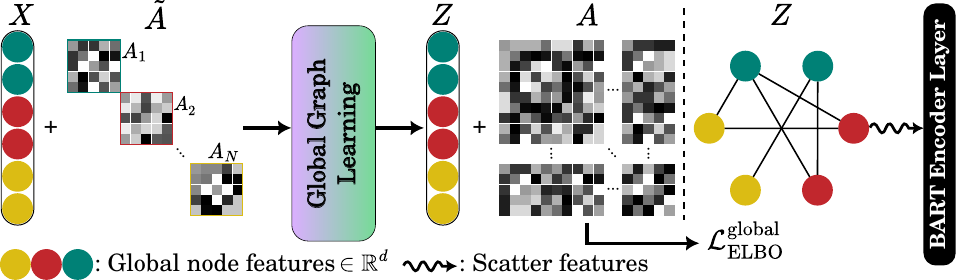}
        }
        \caption{
            We use the predicted local latent graphs $\{A_i\}$ to initialize $\tilde{A} = \mathrm{diag}([A_1,.., A_N], 0)$ in order to learn the final global latent graph $A$. The updated node features $Z$ are scattered back to their initial positions in the BART layer.
        }
    \label{fig:method_stage_2}
  \end{subfigure}
  \hfill
  \begin{subfigure}{0.32\linewidth}
        \centering
        \scalebox{0.7}[0.7]{
        \includegraphics[width=\textwidth]{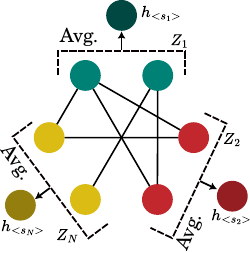}
        }
        \caption{
        We update the state embeddings $h_{<s_i>}$ by averaging the corresponding features from $Z$.
        }
    \label{fig:state_update}
  \end{subfigure}
  \caption{Overview of mixing stage \rom{2}.}
  \label{fig:short}
\end{figure}

\noindent Although the conditional distribution $P(A_i', \tilde{A}_i|X)$ can be modeled by a parametric families of distributions $p^i_\theta(A_i', \tilde{A}_i|X)$, the optimal parameter set $\bar{\theta}$ is not known making the computations of the marginal 
\begin{equation}
    p^i_\theta(\tilde{A}_i|X_i) = \displaystyle \int p^i_\theta(A_i', \tilde{A}_i|X_i) d(A_i')
\end{equation}
and therefore, the posterior of each modality
\begin{equation}
    p^i_\theta(A_i'|\tilde{A}_i, X_i) = \displaystyle \frac{p^i_\theta(A_i', \tilde{A}_i| X_i)}{p^i_\theta(\tilde{A}_i| X_i)}
    \label{eq:mmc}
\end{equation}
intractable.
To be able to infer the missing part of the local adjacency matrix, we take advantage of Variational Inference (VI) to learn an approximation $q^i_\phi(A_i'|\tilde{A}_i, X_i)$ of the posterior. 
We postulate that the missing adjacency matrix of modality $i$ depends on its own features $X_i$ and the features of other modalities $X_{j\neq i}$.
Therefore, we propose a multi-modal conditioning (MMC) 
of \autoref{eq:mmc} on all $X_{j\neq i}$ in addition to $X_i$.
We also follow the idea of \cite{chen2020iterative} that better graph structures lead to better features, and better features lead to better graph structures.
Therefore, as shown in \autoref{fig:method_stage_1},
we use a two-stream approach where one stream uses enhanced features to learn the latent multi-modal graphs, and the other uses the predicted graphs to infer fine-grained features to learn both $q_\phi^i$ and $p_\theta^i$ for each modality.  
Specifically, in the \colorbox{mypurple}{purple} module of the upper stream, we estimate an edge of latent graph $A'_{i,j}$ using cosine similarity as
\begin{equation}\label{eq:graph_inferece}
    a'_{mn} = \frac{1}{K} \sum_{k=1}^K \mathrm{cos}(w_j^k \odot x_m, w_j^k \odot x_n),
\end{equation}
where $x_m, x_n\in X_i$, $\{w_j^k\}$ are learnable weights for each modality, and $\odot$ denotes element-wise multiplication.
Then, in the \colorbox{mygreen}{green} module, we update the multi-modal node features using an APPNP \cite{appnp} module and the predicted latent graphs for modality $i$ to get $\{Z'_{i,j}\}_j$.
For the lower stream, we first start by updating the node features similarly to the upper stream by using the initial graphs $\{\tilde{A}_i\}$ to get $\{Z''_{i,j}\}_j$. 
Then, we use the enhanced node features $\{[Z'_{i, j}, Z''_{i, j}]\}_j$ to predict the second set of local latent graphs $\{A''_{ij}\}_j$.
At the end, we output the final local latent graph of modality $i$ as 
\begin{equation}
   A_i = \underbrace{\frac{1}{2} \tilde{A}_i}_{\textcolor{mygold}{\textrm{\textbf{Initialization Bias (IB)}}}} + \underbrace{\frac{1}{2} \sum_{j=1}^N \frac{1}{N}(A'_{i,j} + A''_{i,j})}_{\textrm{\textcolor{mypurpletext}{\textbf{VI approximation via MMC}}}} \in \mathbb{R}^{K\times K}.
   \label{eq:ib_mmc}
\end{equation}

\subsubsection{Mixing Stage \rom{2} (Conquer).}
This stage tries to infer the global latent graph structure governing the mix of all modalities $\{X_i\}$.
As seen in \autoref{fig:method_stage_2}, it depends on the previously predicted local latent graphs to build the initial global graphs as 
\begin{equation}
    \tilde{A} = \mathrm{diag}([A_1,.., A_N], 0) \in \mathbb{R}^{NK \times NK}.
\end{equation}
Similar to Stage \rom{1}, we use a two-stream approach to learn the global $p_\theta$ and $q_\phi$ and thus the global latent graph $A$ and node features 
\begin{equation}\label{eq:lambda}
    Z = \frac{1}{2}(Z' + Z''),
\end{equation}
where $Z'$ and $Z''$ are obtained from the upper and lower streams, respectively.
Finally, we update the state tokens embeddings $h_{<s_i>}$ by averaging the corresponding features from $Z$ (see \autoref{fig:state_update}) and integrate the latter back into the hidden state of the corresponding BART layer following
\begin{equation}
   H = (1 - \lambda) (H \oslash (Z, \mathrm{Idx})) + \lambda H,
\end{equation}
where $\lambda\in(0,1)$ is a hyper-parameter and $\oslash$, $H$, and $\mathrm{Idx}$ denote the scatter operation, the hidden state of the BART layer and the indices of the nodes features $Z$ relative to $H$, respectively.

\paragraph{Loss Function.}
Since we rely on VI to infer the local and global latent graphs, we used two ELBO losses to optimize (1) the local multi-modal graph learners $\{q_\phi^i, p_\theta^i\}$ and (2) the global learners ${q_\phi, p_\theta}$. 
Please refer to the supplementary material for the derivation of these losses.
We trained our model end-to-end using a combination of the generative loss of the video dialog task $\mathcal{L}_\mathrm{gen}$ and both ELBO losses, i.e. 
\begin{align}
    \mathcal{L} = \alpha_1 \mathcal{L}_\mathrm{gen} &- \alpha_2 \mathcal{L}_\textrm{ELBO}^\mathrm{local} - \alpha_3 \mathcal{L}_\textrm{ELBO}^\mathrm{global}, \label{eq:loss} \\
    \mathcal{L}_\textrm{ELBO}^\mathrm{local} &=  \frac{1}{N}\sum_{i=1}^N  \mathcal{L}_\textrm{ELBO}^{\textrm{local}, i},
\end{align}
where $\{\alpha_k\}$ are hyper-parameters and $\mathcal{L}_\textrm{ELBO}^{\textrm{local}, i}$ is the local ELBO loss for the $i$-th modality.

\section{Experiments}
\subsection{Datasets}

We mainly evaluated our model on the popular and challenging Audio-Visual Scene Aware Dialog (AVSD) dataset \cite{avsd}.
Each of its dialogs comes with $10$ question-answer pairs as well as a short description/caption based on a video.
Each video is collected from the Charades dataset \cite{charades} and the dialogs are generated by human annotators.
We considered all three benchmarks of the dataset, i.e. AVSD-DSTC7 \cite{dstc7}, AVSD-DSTC8 \cite{dstc8}, and AVSD-DSTC10 \cite{av_trn}, which were respectively released for the Dialog System Technology Challenge 
\href{https://github.com/dialogtekgeek/DSTC8-AVSD_official}{(DSTC)}.
To assess the generalizability of our model, we not only experimented with the generative task of SIMMC 2.0 \cite{simmc} but also with the recent and challenging open-ended video question answering NExT-QA dataset \cite{nextqa}.
We refer to the supplementary material for more details about all \textit{five} benchmarks.

\subsection{Metrics}
\label{sec:metrics}
We used the established official metrics for each dataset in order to fairly compare \modelname\, with the previous models.
Specifically, for all \textit{three} AVSD datasets, we used BLEU (B-n) \cite{bleu}, ROUGE-L (R) \cite{rouge}, METEOR (M) \cite{meteor}, and CIDEr (C) \cite{cider}. Whereas for SIMMC and NExT-QA, we used B-4 and WUPS \cite{wups} scores, respectively.

\begin{table}[!t]
    \caption{Results on AVSD-DSTC7 and AVSD-DSTC8.
        Best and second best performances are in \textbf{bold} and \underline{underlined}, respectively.
        $\spadesuit = $ Two-stage training.
        }
    \centering
    \scalebox{0.78}[0.78]{
        
        \begin{tabular}{llcccccccccccccccc}
        \toprule
        \multirow{2}*{\textbf{Model}} & \multirow{2}*{\textbf{Venue}} & & \multicolumn{7}{c}{\textbf{AVSD-DSTC7}} & & \multicolumn{7}{c}{\textbf{AVSD-DSTC8}}\\
        \cmidrule(r){4-10} \cmidrule(r){12-18}
        & & &\textbf{B-1} & \textbf{B-2} & \textbf{B-3} & \textbf{B-4} & \textbf{M} & \textbf{R} & \textbf{C} & &\textbf{B-1} & \textbf{B-2} & \textbf{B-3} & \textbf{B-4} & \textbf{M} & \textbf{R} & \textbf{C} \\
        \midrule
{Baseline} \cite{avsd_baseline} & \textit{ICASSP'19} &
& $62.1$ & $48.0$ & $37.9$ & $30.5$ & $21.7$ & $48.1$ & $73.3$ &
& $61.4$ & $46.7$ & $36.5$ & $28.9$ & $21.0$ & $48.0$ & $65.1$ \\

{MTN} \cite{mtn}                & \textit{ACL'19} &
& $71.5$ & $58.1$ & $47.6$ & $39.2$ & $26.9$ & $55.9$ & $106.6$ &
& $-$ & $-$ & $-$ & $-$ & $-$ & $-$ & $-$ \\
\hline
{JMAN} \cite{chu2020multi}      & \textit{AAAI'20}   &
& $66.7$ & $52.1$ & $41.3$ & $33.4$ & $23.9$ & $53.3$ & $94.1$&
& $64.5$ & $50.4$ & $40.2$ & $32.4$ & $23.2$ & $52.1$ & $87.5$ \\

{VGD} \cite{le-hoi-2020-video}  & \textit{ACL'20} &
& $74.9$ & $62.0$ & $52.0$ & $43.6$ & $28.2$ & $58.2$ & $119.4$ &
& $-$ & $-$ & $-$ & $-$ & $-$ & $-$ & $-$\\

{BiST} \cite{bist}              & \textit{EMNLP'20}  &
& $75.5$ & $61.9$ & $51.0$ & $42.9$ & $28.4$ & $58.1$ & $119.2$ &
& $68.4$ & $54.8$ & $45.7$ & $37.6$ & $27.3$ & $56.3$ & $101.7$ \\
\midrule
{SCGA} \cite{scga}              & \textit{AAAI'21}   &
& $74.5$ & $62.2$ & $51.7$ & $43.0$ & $28.5$ & $57.8$ & $120.1$ &
& $71.1$ & $59.3$ & $49.7$ & $41.6$ & $27.6$ & $56.6$ & $112.3$ \\

{RLM}  \cite{9376902}           & \textit{TASLP'21}  &
& $76.5$ & $64.3$ & $54.3$ & $45.9$ & $29.4$ & $60.6$ & $130.8$ &
& $74.6$ & $62.6$ & $52.8$ & $44.5$ & $28.6$ & $59.8$ & $124.0$ \\

{PDC}  \cite{le2021learning}    & \textit{ICLR'21}   &
& $77.0$ & $65.3$ & $53.9$ & $44.9$ & $29.2$ & $60.6$ & $129.5$ &
& $74.9$ & $62.9$ & $52.8$ & $43.9$ & $28.5$ & $59.2$ & $120.1$ \\
\midrule
{AV-TRN} \cite{av_trn}     & \textit{ICASSP'22}   &
& $-$ & $-$ & $-$ & $40.6$ & $26.2$ & $55.4$ & $107.9$ &
& $-$ & $-$ & $-$ & $39.4$ & $25.0$ & $54.5$ & $99.7$ \\

{VGNMN} \cite{vgnmn}            & \textit{NAACL'22}  &
& $-$    & $-$    & $-$    & $42.9$ & $27.8$ & $57.8$ & $118.8$ &
& $-$ & $-$ & $-$ & $-$ & $-$ & $-$ & $-$ \\ 

{COST} \cite{pham2022video}     & \textit{ECCV'22}   &
& $72.3$ & $58.9$ & $48.3$ & $40.0$ & $26.6$ & $56.1$ & $108.5$ &
& $69.5$ & $55.9$ & $46.5$ & $3.82$ & $27.8$ & $57.4$ & $105.1$ \\

{MRLV} \cite{mrlv}     & \textit{NeurIPS'22}   &
& $-$ & $59.2$ & $49.3$ & $41.5$ & $26.9$ & $56.9$ & $115.9$ &
& $-$ & $-$ & $-$ & $-$ & $-$ & $-$ & $-$ \\

$^{\spadesuit}${THAM} \cite{tham}              & \textit{EMNLP'22}  &
& $77.8$ & $65.4$ & $54.9$ & $46.8$ & \underline{$30.8$} & \underline{$61.9$} & $133.5$ &

& \underline{$76.4$} & \underline{$64.1$} & $53.8$ & $45.5$ & \underline{$30.1$} & \underline{$61.0$} & \underline{$130.4$}
\\
\midrule
{DialogMCF} \cite{dmcf}         & \textit{TASLP'23}  &
& $77.7$ & $65.3$ & $54.7$ & $45.7$ & $30.6$ & $61.3$ & \underline{$135.2$} &
& $75.6$ & $63.3$ & $53.2$ & $44.9$ & $29.3$ & $60.1$ & $125.3$\\

{ITR} \cite{itr}                & \textit{PAMI'23} &
& \underline{$78.2$} & \underline{$65.5$} & \underline{$55.2$} & \underline{$46.9$} & $30.5$ & \underline{$61.9$} & $133.1$  &

& $76.2$ & \underline{$64.1$} & \underline{$54.3$} & \underline{$46.0$} & $29.8$ & $60.7$ & $128.5$
\\
\midrule

\rowcolor{cyan!15} \modelname \parbox[c]{1em}{\includegraphics[width=0.12in]{figures/mixer_icon.pdf}} &      & 
& $\mathbf{78.7}$    & $\mathbf{66.5}$     & $\mathbf{56.3}$     & $\mathbf{47.6}$     & $\mathbf{31.3}$     & $\mathbf{62.5}$     & $\mathbf{138.8}$ &
& $\mathbf{77.5}$     & $\mathbf{66.0}$     & $\mathbf{56.1}$     & $\mathbf{47.7}$    & $\mathbf{30.6}$     & $\mathbf{62.4}$     & $\mathbf{135.4}$\\
\rowcolor{cyan!15} \, w/o $V_\textrm{sam}$ &   \textit{ECCV'24}   & 
& ${78.6}$    & ${66.3}$     & ${56.0}$     & ${47.4}$     & ${31.2}$     & ${62.2}$     & ${137.3}$ &
&${77.4}$    & ${65.8}$     & ${56.0}$     & ${47.3}$     & ${30.6}$     & ${62.1}$     & ${134.8}$\\
\rowcolor{cyan!15} \quad w/o $A_\textrm{vggish}$ &      & 
& ${78.4}$    & ${66.0}$     & ${55.8}$     & ${47.1}$     & ${31.0}$     & ${62.0}$     & ${136.5}$ &
& $77.1$    & ${65.6}$     & ${55.7}$     & ${47.1}$     & ${30.2}$     & ${61.8}$     & ${133.6}$  \\

\bottomrule                                            
\end{tabular}
    }
    
    \label{tab:dstc78}
\end{table}

\begin{table}[!t]
    \caption{Results on AVSD-DSTC10.}
    \centering
    \scalebox{0.9}[0.9]{
        \begin{tabular}{llccccccc}
        \toprule
        \textbf{Model} & \textbf{Venue} & \textbf{B-1} 
        & \textbf{B-2} & \textbf{B-3} & \textbf{B-4} & \textbf{M} & \textbf{R} & \textbf{C} \\
        \midrule
            AV-TRN \cite{av_trn} & \textit{ICASSP'22} & $-$ & $-$ & $-$ & $24.7$ & $19.1$ & $43.7$ & $56.6$ \\
            \quad + Ext. \cite{av_trn} & \textit{ICASSP'22} & $-$ & $-$ & $-$ & $37.1$ & $24.5$ & $53.5$ & $86.9$  \\
            DSTC10 \cite{best_dstc10}   & \textit{AAAI'22} & $67.3$ & $54.5$     & $44.8$  & \underline{$37.2$} & $24.3$ & $53.0$ & \underline{$91.2$} \\
            DialogMCF  \cite{dmcf} & \textit{TASLP'23} & \underline{$69.3$} & \underline{$55.6$} & \underline{$45.0$} & $36.9$ & \underline{$24.9$} & \underline{$53.6$} & \underline{$91.2$} \\
            \midrule
            \rowcolor{cyan!15} \modelname \parbox[c]{1em}{\includegraphics[width=0.12in]{figures/mixer_icon.pdf}}  &  &
            $\mathbf{70.0}$     & $\mathbf{57.4}$     & $\mathbf{47.6}$     & $\mathbf{40.0}$     & $\mathbf{25.7}$     & $\mathbf{54.5}$     & $\mathbf{99.8}$ \\
            \rowcolor{cyan!15} \, w/o $V_\textrm{sam}$  & \textit{ECCV'24}  &
            ${69.8}$     & ${57.4}$     & ${47.5}$     & ${39.8}$     & ${25.6}$     & ${54.3}$     & ${97.6}$ \\
            \rowcolor{cyan!15} \quad w/o $A_\textrm{vggish}$ &  &
            ${69.7}$     & ${57.1}$     & ${47.2}$     & ${39.5}$     & ${25.1}$     & ${54.0}$     & ${96.9}$ \\
            \bottomrule
        \end{tabular}
    }
    \label{tab:avsd_10}
\end{table}

\begin{table}[!t]
\begin{minipage}[b]{.4\textwidth}
    \caption{Results on SIMMC.}
        \centering
         \scalebox{0.8}[0.8]{

            \begin{tabular}{llc}
                \toprule
                \multirow{1}*{\textbf{Model}} & \textbf{Venue} & \textbf{B-4} \\
                \midrule
                {MTN} \cite{mtn} & \textit{ACL'19} &$21.7$  \\ 
                {GPT-2} \cite{simmc} & \textit{EMNLP'21} & $19.2$ \\ 
                {BART} \cite{lee-etal-2022-learning} & \textit{NAACL'22} & $33.1$\\ 
                {PaCE} \cite{pace} & \textit{ACL'23} &\underline{$34.1$} \\ 
                \midrule
                \rowcolor{cyan!15}  \modelname \parbox[c]{1em}{\includegraphics[width=0.12in]{figures/mixer_icon.pdf}} & \textit{ECCV'24} &$\mathbf{44.7}$\\ 
                \bottomrule
            \end{tabular}
        }
        \label{tab:simmc}
\end{minipage}
\begin{minipage}[b]{.6\textwidth}
    \caption{Results on open-ended NExT-QA$^\diamondsuit$.}
    \centering
    \scalebox{0.7}[0.7]{
        
        \begin{tabular}{llcccc}
        \toprule
        \textbf{Model} & \textbf{Venue} & \textbf{WUPS}$_C$ 
        & \textbf{WUPS}$_T$ & \textbf{WUPS}$_D$ & \textbf{WUPS} \\
        \midrule
            {HCRN} \cite{hcrn} & \textit{CVPR'20} & $16.05$ & $17.68$ & $49.78$ & $23.92$  \\
            {HGA} \cite{hga} & \textit{AAAI'20} & \underline{$17.98$} & \underline{$17.95$} & \underline{$50.84$} & $24.06$  \\
            {Flamingo} \cite{flamingo}   & \textit{NeurIPS'22} & $-$ & $-$ & $-$ & \underline{$28.40$}  \\
            
            {KcGA} \cite{kcga} & \textit{AAAI'23} & $-$ & $-$ & $-$ & $28.20$  \\
            {EMU} \cite{emu}  & \textit{arXiv'23} & $-$ & $-$ & $-$ & $23.40$  \\

            \midrule
            \rowcolor{cyan!15} \modelname \parbox[c]{1em}{\includegraphics[width=0.12in]{figures/mixer_icon.pdf}} & \textit{ECCV'24} &
            $\mathbf{22.12}$     & $\mathbf{22.20}$     & $\mathbf{55.64}$     & $\mathbf{29.50}$     \\
            \bottomrule
            
        \end{tabular}
    }    
    \label{tab:nextqa}
\end{minipage}%
\end{table}

\subsection{Main Results}

\subsubsection{AVSD-DSTC7.}

As can be seen in \autoref{tab:dstc78}, our model managed to achieve new SOTA results across all evaluation metrics, thereby outperforming the latest baselines, including PDC \cite{le2021learning}, DialogMCF \cite{dmcf}, THAM \cite{tham}, and ITR \cite{itr}.
Specifically, \modelname\, outperformed the latest ITR \cite{itr} model by over 1.5\% (relative improvement) on B-2, B-3, B-4, and M scores.
Since some previous models did not use SAM \cite{sam} and audio features, we trained two additional versions of our model where we only removed SAM features before additionally removing the audio features. Both versions are denoted by ``w/o $V_\textrm{sam}$'' and ``w/o $A_\textrm{vggish}$'', respectively.
As seen from \autoref{tab:dstc78}, both versions still outperform all previous models across all evaluation metrics. 

\subsubsection{AVSD-DSTC8.}\blfootnote{$^\diamondsuit$ C, T, and D denote causal, temporal, and descriptive questions, respectively.}
As depicted in \autoref{tab:dstc78}, models tend to struggle more on this more recent benchmark.
However, \modelname\, scored new SOTA results with higher relative improvements compared to DSTC7, thereby lifting the B-2, B-3, B-4, and C scores by over 3\% relative to the second best models ITR \cite{itr} and THAM \cite{tham}.
Similarly to AVSD-DSTC7, our ablated versions surpassed these models on all evaluation metrics and marginally underperformed our full model.

\subsubsection{AVSD-DSTC10.}
We then evaluated \modelname\, on the latest AVSD-DSTC10 benchmark.
Contrary to the previous versions, AVSD-DSTC10 does not include human-generated video descriptions during inference since these are unavailable in real-world applications. 
As depicted in \autoref{tab:avsd_10}, models struggle the most on this challenge version.
However, not only our full \modelname\, model but also its two ablated versions managed to outperform the latest models on all evaluation metrics. 

\subsubsection{$^\clubsuit$SIMMC.}\blfootnote{$\clubsuit$: Models trained with optimal hyperparameters from AVSD and without $V_\textrm{sam}$.}
To assess the generalizability of our model, we additionally tested it on the generative task of SIMMC 2.0 \cite{simmc0}. 
As seen from \autoref{tab:simmc}, \modelname\, outperformed the latest published models such as PaCE \cite{pace} by achieving a B-4 score of $44.7$.

\subsubsection{$^\clubsuit$NExT-QA.}
Finally, we tested our model on the recent open-ended NExT-QA benchmark \cite{nextqa}.
As depicted in \autoref{tab:nextqa}, \modelname\, not only outperformed HCRN \cite{hcrn} and HGA \cite{hga} on all WUPS scores \cite{wups} but also surpassed latest models such as Flamingo \cite{flamingo}, KcGA \cite{kcga}, and EMU \cite{emu}. 
Specifically, it lifted the overall WUPS score by 1.1 absolute points compared to the seminal Flamingo-9B model with x18 more parameters.

\subsection{Ablation Study}
\subsubsection{Effect of $\lambda$ and $\Delta$.}
We independently optimized these hyper-parameters based on the validation perplexity (PPL).
First, we fixed $\Delta=4$ to guarantee a reasonable training time on our hardware setup and varied $\lambda \in \{0, 0.1, 0.5, 0.9, 1\}$.
As seen in \autoref{tab:lambda}, the best performance was achieved when using $\lambda=0.9$.
Thereafter, we varied $\Delta \in \{2, 3, 4, 5\}$ while keeping $\lambda=0.9$ and achieved the best results for $\Delta=4$ as can be seen from \autoref{tab:delta}.

\begin{table}[!t]
\begin{minipage}[b]{.5\textwidth}
    \caption{Influence of the value of $\lambda$.}
    \centering
    \scalebox{0.73}[0.73]{
        
        \begin{tabular}{cccccccc}
        \toprule
          \multirow{2}*{$\lambda$} & \multirow{2}*{\textbf{PPL}} & \multicolumn{3}{c}{\textbf{AVSD-DSTC7}} & \multicolumn{3}{c}{\textbf{AVSD-DSTC8}} \\
          
          \cmidrule(r){3-5}   \cmidrule(r){6-8}

        & \textbf{(val)} & \textbf{B-4} & \textbf{R} & \textbf{C} & \textbf{B-4} & \textbf{R} & \textbf{C} \\
        \midrule
        $0.0$  & 
        \multicolumn{7}{c}
        {\cellcolor{red!20}\texttt{Training unstable}}\\
        $0.1$& $11.03$ & $17.3$  & $29.0$  & $35.1$ & $11.4$ & $24.3$ & $21.2$\\
        $0.5$& $5.48$  & $44.6$  & $60.3$  & $126.4$ & $44.7$ & $59.4$ & $123.8$\\
        $0.9$& $\mathbf{5.16}$ & $\mathbf{47.6}$ & $\mathbf{62.5}$  & $\mathbf{138.8}$  & $\mathbf{47.7}$  & $\mathbf{62.4}$ & $\mathbf{135.4}$\\
        $1.0$& ${5.30}$ & ${45.1}$  & ${60.8}$  & ${131.3}$  & ${42.3}$ & ${61.1}$ & ${126.9}$\\

        \bottomrule
        \end{tabular}
    }

    \label{tab:lambda}
\end{minipage}
\begin{minipage}[b]{.5\textwidth}
    \caption{Influence of the value of $\Delta$.}
    \centering
        \scalebox{0.8}[0.8]{
        
        \begin{tabular}{cccccccc}
        \toprule
          \multirow{2}*{$\Delta$} & \multirow{2}*{\textbf{PPL}} & \multicolumn{3}{c}{\textbf{AVSD-DSTC7}} & \multicolumn{3}{c}{\textbf{AVSD-DSTC8}} \\
          
          \cmidrule(r){3-5}   \cmidrule(r){6-8}

        & \textbf{(val)} & \textbf{B-4} & \textbf{R} & \textbf{C} & \textbf{B-4} & \textbf{R} & \textbf{C} \\
        \midrule
        $\leq 2$  & \multicolumn{7}{c}
        {\cellcolor{red!20}\texttt{Training too long}}
        \\
        $3$& $5.19$ & $45.7$  & $61.5$  & $134.1$ & $46.7$ & $61.5$ & $131.8$\\
        $4$& $\mathbf{5.16}$ & $\mathbf{47.6}$ & $\mathbf{62.5}$  & $\mathbf{138.8}$  & $\mathbf{47.7}$  & $\mathbf{62.4}$ & $\mathbf{135.4}$\\

        $5$& $5.21$ & $45.0$  & $61.1$  & $133.6$ & $44.6$ & $60.5$ & $129.1$\\

        \bottomrule
        \end{tabular}
    }
    \label{tab:delta}
\end{minipage}%
\end{table}

\subsubsection{Latent Graph Size $K$.}
\begin{figure}[!t]
    \begin{minipage}{1\linewidth}
        \centering
        \scalebox{1}[1]{
            \includegraphics[width=\textwidth]{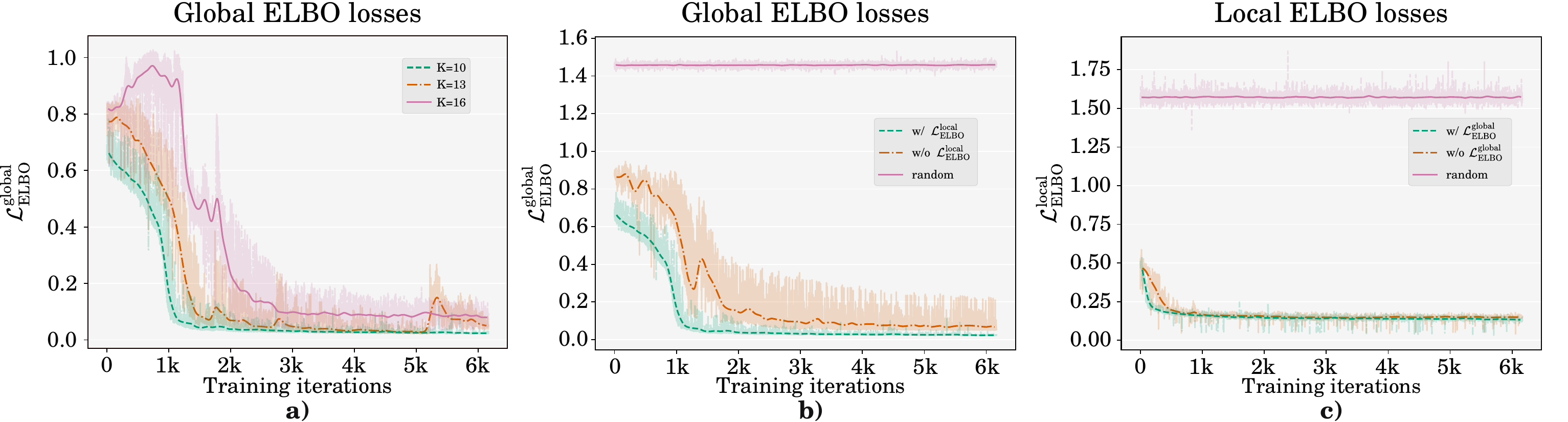}
        }
        \caption{\textbf{a)} Larger values of $K$ make the learning of the global latent graphs more challenging.
        \textbf{b)} The local ELBO loss $\mathbf{\mathcal{L}_\textrm{ELBO}^\mathrm{local}}$ facilitates the learning of the global latent graphs.
        \textbf{c)} The global ELBO loss $\mathbf{\mathcal{L}_\textrm{ELBO}^\mathrm{global}}$ facilitates the learning of the local latent graphs.
        All models use SAM and audio features.
        }
        \label{fig:elbos}
    \end{minipage}
\end{figure}

As illustrated in the first section of \autoref{tab:ablation_1}, we varied $K$ from $7$ to $16$ in three-step intervals.
The overall performance of \modelname\, peaked when using $K=10$ tokens from each modality as the graphs' node features.
Using higher values of $K$ rendered the learning of the global latent graphs with $K\times N$ nodes more difficult and thus hurt the overall performance of our model.
This is underlined by the behavior of the global ELBO loss $\mathcal{L}_\textrm{ELBO}^\mathrm{global}$ as illustrated in \autoref{fig:elbos}\textcolor{red}{a}.
Using $K=7$ hurt the performance of our model across almost all metrics.
We posit that low values of $K$ are insufficient to capture each modality's most influential constituents. 
Therefore, we set $K=10$ in the rest of the experiments.

\subsubsection{Multi-Modal State Tracking GNNs.}
In each row of the middle section of \autoref{tab:ablation_1}, we ablated \textit{one} GNN-based tracking module and kept the remaining ones unchanged. 
Our full model outperformed all these ablated versions \textit{despite them having access to the same input features}.
The comparable results of all these ablated versions validate using a uniform graph size $K$ for all different modalities.
Finally, we replaced all GNNs (local and global) with vanilla transformer layers.
As can be seen from the \colorbox{orange!25}{last row} of the middle section, this version was outperformed by our full model as well, underlining the efficacy of our proposed multi-modal graph learning approach.  

\begin{table}[!t]
    \caption{Comparison between different ablated versions of our model. 
    All ablations use SAM and audio features.
    \colorbox{orange!25}{TRN} means that the model replaces the global and local multi-modal GNNs with vanilla transformer layers, and \colorbox{red!20}{RAND} denotes that it uses random latent graphs instead of learning them.
    Our full model is highlighted in \colorbox{cyan!15}{blue}.
    }
    
    \centering
    \scalebox{0.9}[0.9]{
        
        \begin{tabular}{cccccccccccccc}
        \toprule
          \multirow{2}*{$\mathbf{K}$} & \multirow{2}*{\textbf{GNNs}} & \multirow{2}*{$\mathbf{\mathcal{L}_\textrm{ELBO}^\mathrm{local}}$} & \multirow{2}*{$\mathbf{\mathcal{L}_\textrm{ELBO}^\mathrm{global}}$} & \multirow{2}*{$\mathbf{\#}$ \textbf{Params.}} & \multicolumn{4}{c}{\textbf{AVSD-DSTC7}} & \multicolumn{4}{c}{\textbf{AVSD-DSTC8}} \\
          \cmidrule(r){6-9}   \cmidrule(r){10-13}

        &  & & & &  \textbf{B-1} 
        & \textbf{B-4} & \textbf{R} & \textbf{C} & \textbf{B-1} 
        & \textbf{B-4} & \textbf{R} & \textbf{C} \\
        \midrule
        $ 7$ & All & \cmark & \cmark & $\sim 511$M &$77.8$  & $47.0$  & $61.8$  & $136.2$  & $76.6$  & $47.0$  & $61.5$  & $131.8$ \\
        \rowcolor{cyan!15}
        $10$ & All & \cmark & \cmark & $\sim 511$M & $\mathbf{78.7}$  & $\mathbf{47.6}$  & $\mathbf{62.5}$  & $\mathbf{138.8}$  & $\mathbf{77.5}$  & $\mathbf{47.7}$  & $\mathbf{62.4}$  & $\mathbf{135.4}$ \\
        $13$ & All & \cmark & \cmark & $\sim 511$M & $77.0$  & $45.4$  & $60.6$  & $131.9$  & $75.7$  & $45.2$  & $60.4$  & $127.0$ \\
        $16$ & All & \cmark & \cmark & $\sim 511$M & $76.6$  & $45.4$  & $60.7$  & $132.6$  & $75.8$  & $45.9$  & $60.5$  & $128.4$ \\
        \midrule
        $10$ & \multicolumn{1}{l}{w/o GNN$_\textrm{rgb}$} & \cmark & \cmark & $\sim 495$M & $78.4$  & \underline{$47.2$}  & $62.4$  & $137.2$ &  $77.3$  & \underline{$47.4$}  & $62.0$  & $133.2$ \\
        $10$ & \multicolumn{1}{l}{w/o GNN$_\textrm{flow}$} & \cmark & \cmark  & $\sim 495$M & \underline{$78.5$}  & $47.1$  & \underline{$62.5$}  & \underline{$138.5$} &  $76.9$  & $47.2$  & $61.9$  & $\underline{134.1}$ \\
        $10$ & \multicolumn{1}{l}{w/o GNN$_\textrm{sam}$} & \cmark & \cmark & $\sim 495$M & $78.1$  & $46.1$  & $62.2$  & $137.2$ &  \underline{$77.5$}  & $46.5$  & $61.7$  & $132.7$ \\
        $10$ & \multicolumn{1}{l}{w/o GNN$_\textrm{vggish}$} & \cmark & \cmark & $\sim 495$M & $78.0$ & $45.8$  & $61.4$  & $134.9$ &  $76.8$  & $46.5$  & $61.0$  & $131.0$ \\
        $10$ & \multicolumn{1}{l}{w/o GNN$_\textrm{H}$} & \cmark & \cmark & $\sim 495$M & $78.1$  & $45.7$  & $61.8$  & $134.1$ &  $77.4$  & $46.7$  & \underline{$62.2$}  & $134.0$ \\
        $10$ & \multicolumn{1}{l}{w/o GNN$_\textrm{Q}$} & \cmark & \cmark & $\sim 495$M & $78.2$  & $47.1$  & $62.1$  & $138.5$ &  $77.0$  & $47.0$  & $61.8$  & $133.6$ \\
        \rowcolor{orange!25} $10$  & TRN & \xmark & \xmark & $\sim 500$M  & $77.8$  & $46.9$  & $61.8$  & $136.6$ &  $76.8$  & $46.7$  & $61.4$  & $131.8$ \\
        \midrule
        $-$  & $-$ & \xmark & \xmark & $\sim 411$M & $76.6$  & $45.1$  & $60.8$  & $131.3$ &  $74.2$  & $42.3$  & $61.1$  & $126.9$ \\
        $-$  & w/ only $\tilde{A}_i$ & \xmark & \xmark & $\sim 413$M & $76.5$    & ${45.4}$     & ${60.9}$  & ${131.7}$ &
        ${75.2}$  & ${45.5}$   & ${60.7}$  & ${130.3}$ & \\

        $10$ & All & \xmark & \cmark & $\sim 416$M & $75.9$  & $44.5$  & $59.8$  & $127.8$  & $74.3$  & $44.2$  &  $59.2$  & $122.8$ \\
        $10$ & All & \cmark & \xmark & $\sim 506$M & $77.5$ & $46.4$  & $61.4$ & $134.9$  & $76.2$ & $46.6$  & $60.9$ & $130.6$  \\
        
        \rowcolor{red!20} $10$  & All & RAND & RAND & $\sim 448$M  & $73.0$  & $42.1$  & $57.3$  & $119.2$ &  $71.4$  & $41.6$ & $57.1$  & $114.2$ \\
        
        \bottomrule
        \end{tabular}
    }

    \label{tab:ablation_1}
\end{table}

\begin{table}[!t]
    \caption{Comparison between different ablated versions of our model. All ablations were trained with SAM and audio features and with the optimal hyper-parameters as the full model. \textcolor{mygold}{\textbf{IB}} = Initialization Bias, \textcolor{mypurpletext}{\textbf{MMC}} = Multi-Modal Conditioning.
    }
    \centering
    \scalebox{0.78}[0.78]{
        
        \begin{tabular}{lcccccccccc}
        \toprule
          \multirow{2}*{\modelname \parbox[c]{1em}{\includegraphics[width=0.12in]{figures/mixer_icon.pdf}}} & \multirow{2}*{$\mathbf{\#}$ \textbf{Params.}} & \multicolumn{4}{c}{\textbf{AVSD-DSTC7}} & \multicolumn{4}{c}{\textbf{AVSD-DSTC8}} \\
          \cmidrule(r){3-6}   \cmidrule(r){7-10}

        & & \textbf{B-1} & \textbf{B-4} & \textbf{R} & \textbf{C} & \textbf{B-1} 
        & \textbf{B-4} & \textbf{R} & \textbf{C} \\
        \midrule
        \textcolor{mypurpletext}{\textbf{w/o MMC}} & $\sim 500$M &
        $76.9$    & ${46.6}$     & ${61.4}$  & ${135.5}$ &
        ${75.8}$  & ${46.1}$  & ${60.5}$     & ${130.9}$ \\
        \textcolor{mygold}{\textbf{w/o IB}}  & $\sim 511$M &
        $77.6$    & ${47.0}$     & ${61.8}$  & ${136.2}$ &
        ${76.3}$  & ${46.2}$     & ${61.2}$  & ${131.1}$\\
        \midrule
        \rowcolor{cyan!15} \textbf{Full} & $\sim 511$M & $\mathbf{78.7}$  & $\mathbf{47.6}$  & $\mathbf{62.5}$  & $\mathbf{138.8}$  & $\mathbf{77.5}$  & $\mathbf{47.7}$  & $\mathbf{62.4}$  & $\mathbf{135.4}$ \\
        \bottomrule
        \end{tabular}
    }

    \label{tab:ablation_last}
\end{table}

\subsubsection{ELBO Losses.}
As can be seen in the third section of \autoref{tab:ablation_1}, we conducted extensive experiments with different combinations of the ELBO losses:
(1) We first ablated the learning of both global and local latent graphs and, therefore, both ELBO losses resulting in a plain BART model \cite{Lewis2020}.
(2) We then only used the initial graphs $\tilde{A}_i$ as the final latent graph approximations in both training stages \rom{1} and \rom{2} leading to improvements compared to plain BART.
(3) Thereafter, we ablated the local ELBO loss and directly learned the global latent graphs.
This version of our model underperformed BART, which follows our hypothesis that directly learning the global latent graphs is daunting and might lead to performance drops.
As illustrated in \autoref{fig:elbos}\textcolor{red}{b}, $\mathbf{\mathcal{L}_\textrm{ELBO}^\mathrm{global}}$ converged faster and reached lower values when optimized jointly with $\mathbf{\mathcal{L}_\textrm{ELBO}^\mathrm{local}}$.
(4) We thereafter ablated the global ELBO loss and only learned the local latent graphs, leading to performance increases compared to the previous versions.
This underlines that learning the local latent graphs is less sensitive to $\mathbf{\mathcal{L}_\textrm{ELBO}^\mathrm{global}}$ than learning the global latent graphs is to $\mathbf{\mathcal{L}_\textrm{ELBO}^\mathrm{local}}$ as can be seen in \autoref{fig:elbos}\textcolor{red}{c}.     
(5) We finally evaluated a version with a comparable computational complexity as our full model but used random latent graphs instead of learning them.
As can be seen in \autoref{fig:elbos}\textcolor{red}{b}, \autoref{fig:elbos}\textcolor{red}{c}, and the \colorbox{red!20}{last row} of \autoref{tab:ablation_1}), both ELBO losses remained constant and the model reached the worst results among all ablated versions empirically showcasing the importance of our latent graph learning approach.

\subsubsection{Latent Graph Learning.}
Lastly, we considered two additional ablations of \modelname.
Specifically, we first ablated the multi-modal conditioning (MMC) of \autoref{eq:mmc} and learned the local latent graphs of modality $i$ based only on its features $X_i$. This reduces \autoref{eq:ib_mmc} to 
\begin{equation}
   A_i = \frac{1}{2} \tilde{A}_i + \frac{1}{2} (A'_{i} + A''_{i}).
\end{equation}
Then, we trained a version without the initialization bias (IB) of \autoref{eq:ib_mmc}.
As can be seen in \autoref{tab:ablation_last}, MMC is essential for high performance. Without it \modelname\, achieved the lowest performance across all metrics.
The same applies to IB since not incorporating $\tilde{A}_i$ and only using the posterior approximation impeded the performance across all evaluation metrics.  

\begin{figure*}[!t]
    \begin{minipage}{1\linewidth}
        \centering
        \scalebox{0.93}[0.93]{
            \includegraphics[width=\textwidth]{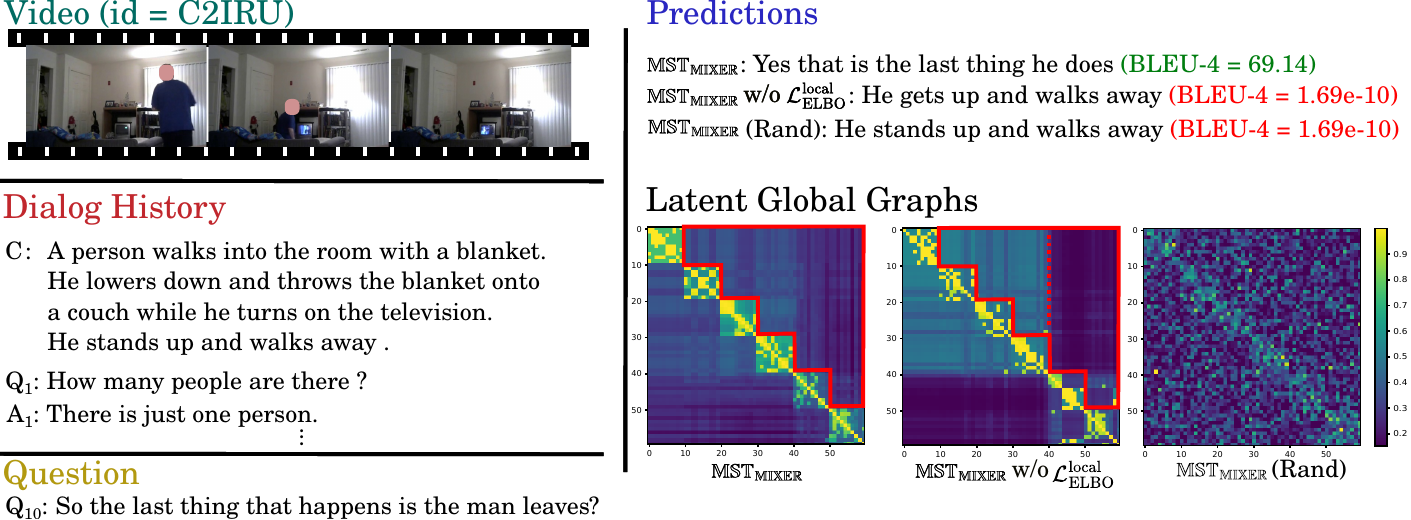}
        }
        \caption{Qualitative comparison of different model ablations.
        on response generation and latent global graph inference of $q_\phi$ obtained from the last encoder layer.
        The diagonal blocks 
        (from upper left to lower right)
        correspond to $V_\textrm{rgb}, V_\textrm{flow}, V_\textrm{sam}, A_\textrm{vggish}, T_\textrm{H}, \textrm{ and } T_\textrm{Q}$, respectively.
    }
        \label{fig:qualitative}
    \end{minipage}
\end{figure*}

\subsection{Qualitative Results}
Finally, in \autoref{fig:qualitative} we give a qualitative comparison of \modelname\, with different ablated versions on response generation and global latent graph inference: Our full model managed to accurately answer the question whereas both ablated version failed to generate reliable responses.
Furthermore, we can see how our full model better captured the local interactions within each modality (more structured diagonal blocks) as well as the global ones across modalities: Whereas the off-diagonal region (bordered in red) of the version ``w/o $\mathcal{L}_\textrm{ELBO}^\mathrm{local}$'' showed a clear divide between the modalities (dotted line), the full model mitigated this by producing more homogeneous values indicating better inter-modal interactions.
We provide more examples and failure cases in the supplementary material.

\section{Conclusion}
We proposed \modelname-- a novel multi-modal state tracking model specifically geared towards video dialog.
\modelname\ first identifies the most influential constituents at different semantic levels (e.g., across modalities and encoder layers).
Then, it relies on a two-stage divide and conquer approach to infer the missing underlying structure of the mix of all modalities and leverages it to augment the hidden states of the backbone VLM using GNNs.
Through extensive ablations experiments and evaluations on \textit{five} video-and-language benchmarks, we show our approach's effectiveness and generalization capabilities.
\\
\textbf{Acknowledgments.}
L. Shi was funded by the Deutsche Forschungsgemeinschaft (DFG, German Research Foundation) under Germany’s Excellence Strategy -- EXC 2075–390740016.

\appendix
\lstset{style=mystyle}

\section{ELBO Derivation \& Implementation }
\label{appendix:elbos}
In this section, we derive the ELBO loss and show how it can be used as an optimization term in our total loss.
Without the loss of generality, we only consider the ELBO in the global setting.
Given the intractable posterior $p_\theta(A'|\tilde{A}, X)$ and the its approximation $q_\phi(A'|\tilde{A}, X)$, it holds that

\begin{align}
       & \mathcal{D}_\mathrm{KL}\left(q_\phi(A'|\tilde{A}, X) || p_\theta(A'|\tilde{A}, X) \right)
       = \mathbb{E}_{q_\phi(A'|\tilde{A}, X)}\left[ \mathrm{log} \frac{q_\phi(A'|\tilde{A}, X)}{ p_\theta(A'|\tilde{A}, X)} \right] \\
       & = \mathbb{E}_{q_\phi(A'|\tilde{A}, X)}\left[ \mathrm{log} \frac{q_\phi(A'|\tilde{A}, X) p_\theta(\tilde{A}| X) }{ p_\theta(A', \tilde{A} | X)} \right] \\
       & = \mathbb{E}_{q_\phi(A'|\tilde{A}, X)}\left[ \mathrm{log} \frac{q_\phi(A'|\tilde{A}, X)}{ p_\theta(A', \tilde{A} | X)} \right] + p_\theta(\tilde{A}| X)\\
       & =   \underbrace{p_\theta(\tilde{A}| X)}_{\textrm{Evidence}} - \underbrace{\mathbb{E}_{q_\phi(A'|\tilde{A}, X)}\left[ \mathrm{log} \frac{p_\theta(A', \tilde{A} | X)}{q_\phi(A'|\tilde{A}, X)} \right]}_{=: 
        \mathcal{L}_\mathrm{ELBO}^\mathrm{global}} \geq 0
\end{align}

Thus, as its name suggests, ELBO serves as a lower bound of the evidence.
As a results, VI tries to maximize the ELBO which is equivalent to minimizing the Kullback-Leibner Divergence between $q_\phi(A'|\tilde{A}, X)$ and the intractable posterior $p_\theta(A'|\tilde{A}, X)$ leading to better estimation of the latter.
Since we used the ELBOs as terms in the total loss $\mathcal{L}$ to be minimized, we had to use the opposite value of each one of them.
This explains the minus sign in \textcolor{red}{Equation 12} in the main text. 
Since $q_\phi$ and $p_\theta$ only output normalized scores as the prediction for each edge, we appended the zero vectors to both predictions in order to convert the raw scores to a two-value probability before applying the log-softmax function.
We provide in \autoref{lst:elbo} a code-snippet of our implementation of the ELBO loss.

\section{Generative Loss}
In addition to the ELBO losses, we used the generative loss $L_{gen}$ to train our model. 
It employs teacher forcing and teaches the BART decoder to predict the next response token $\hat{y}_{j+1}$ conditioned on the previous ground-truth response tokens $Y_{j} = [{y}_{1}, ..., {y}_{j}]$ and the output of the encoder $H_{enc}$.
Specifically, the next predicted token satisfies 
\begin{equation}
        \hat{y}_{j+1} = \displaystyle \argmax_{y\in V} \left[log P\left(y | {Y}_{j}, H_{enc}\right)\right],%
\end{equation}
where $V$ and $P$ denote the vocabulary and the softmax of the logits of the last decoder layer, respectively.

\section{Datasets}
\label{appendix:dataset}
\subsection{AVSD}
The AVSD dataset \cite{avsd} was released 
in the $7$th Dialogue System Technology Challenge (DSTC7) \cite{dstc7}.
As can be seen from \autoref{tab:avsd_stat}, it contains $7,659$, $1,787$, and $1,710$ dialogs for training, validation and testing, respectively.
The data for DSTC8 \cite{dstc8} and DSTC10 \cite{av_trn} were only released with $1,710$ and $1,804$ dialogs for testing, respectively.
For all testing splits, six human-generated reference answers were provided for each dialog in order to compute the generation metrics.

\subsection{SIMMC2.0}
\begin{figure}[!t]
    \begin{minipage}{1\linewidth}
        \centering
        \caption{
        Comparison between the visual complexity of AVSD \textbf{(a)} and SIMMC 2.0 \textbf{(b)}.
        For ethical reasons, we blurred the faces of people appearing in the video frames.
        }
        \scalebox{0.97}[0.97]{
            \includegraphics[width=\textwidth]{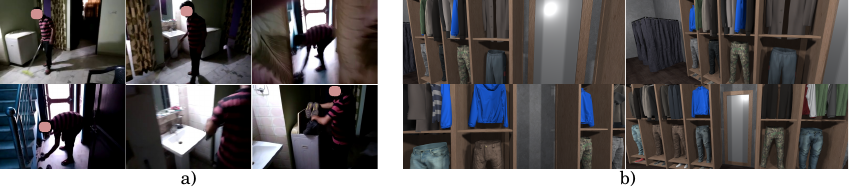}
        }
        \label{fig:comp_avsd_simmc}
    \end{minipage}
\end{figure}

SIMMC 2.0 \cite{simmc} is a task-oriented dataset that was proposed for virtual assistance scenarios and contains $11$k dialogs with $52,044$ unique questions grounded in $5,440$ videos from the shopping domain.
Its visual and textual data were automatically generated in constrained and pre-defined settings resulting in less complex and challenging scenes compared to AVSD. 
As can be seen in \autoref{fig:comp_avsd_simmc}, AVSD features a larger variety of
objects that humans interact with daily, more complex dynamics, and more challenging illumination conditions.
On the other hand, SIMMC 2.0 only comes with simple items linked to the shopping domain.

\begin{table}[!t]
    \centering
    \caption{Summary of the AVSD dataset with all test splits from DSTC7, DSTC8, and DSTC10.
    }
    \label{tab:avsd_stat}
    \scalebox{1}[1]{
        \begin{tabular}{lccccc}
        \toprule
           &  \multirow{2}*{\textbf{Train}} & \multirow{2}*{\textbf{Val}} & \multicolumn{3}{c}{\textbf{Test}} \\
                     \cmidrule(r){4-6}
         & & &  \textbf{DSTC7} & \textbf{DSTC8} & \textbf{DSTC10} \\
        \midrule
        \textbf{\# Dialogs/Videos}     & $7,659$  & $1,787$  & $1,710$  & $1,710$ & $1,804$ \\
        \textbf{\# Questions/Answers}  & $153,180$ & $35,740$  & $13,490$  & $18,810$ & $28,406$\\
        \textbf{\# Words}              & $1,450,754$  & $339,006$  & $110,252$  & $162,226$ & $272,606$ \\
        \bottomrule
        \end{tabular}
    }

\end{table}

\begin{table}[!t]
    \centering
    \caption{Summary of the open-ended NExT-QA dataset.
    }
    \label{tab:nextqa_stat}
    \scalebox{1}[1]{
        
        \begin{tabular}{lccc}
        \toprule
           & \textbf{Train} & \textbf{Val} & \textbf{Test} \\
        \midrule
        \textbf{\# Videos}     & $3,870$  & $570$  & $1,000$ \\
        \textbf{\# Questions}  & $37,523$ & $5,343$  & $9,178$  \\
        \bottomrule
        \end{tabular}
    }

\end{table}

\subsection{NExT-QA}
NExT-QA \cite{nextqa} was recently introduced as a next generation video question answering benchmark that was introduced to advance video
understanding from describing to explaining the temporal actions. 
\autoref{tab:nextqa_stat} gives more insight about the statistics of the dataset. 

\section{Experimental Setup}
\label{appendix:exp_setup}
\subsection{Hardware \& Environment}
We implemented our model in PyTorch \cite{pytorch} and trained them on a cluster consisting of $8$ Nvidia Tesla V100 (32GB) GPUs, $2$ Intel(R) Xeon(R) Platinum 8160 CPUs, and $1.5$TB of RAM.

\subsection{Training}
We trained \modelname\, end-to-end using AdamW \cite{adamw} with $\beta_1 = 0.9$, $\beta_2 = 0.999$, and $\epsilon=1e-8$ and a linear learning rate schedule with warm-up for a maximum of $12$ epochs. 
We utilized a learning rate $\mathrm{lr}_{\mathrm{BART}} = 1e-5$ for the weights of the BART model and a learning rate $\mathrm{lr}_{\mathrm{rest}} = 1e-4$ for the rest of the parameters of our model.
Similarly to $\lambda$ and $\Delta$, we validated the choice of the ELBO loss coefficients $\alpha_2$ and $\alpha_3$ based on the validation perplexity.
Specifically, we performed a grid search using the value set $\{1, 10, 100, 1000\}$ while keeping $\lambda=0.9, \Delta=4$, and $K=10$.
The training of our full model takes approximately $20$ hours to finish.
Complete details about the hyperparameter values
are listed in \autoref{tab:hyperparams_1}.

\subsection{Inference}
Similar to previous works, we utilized beam search with a depth of $5$ and a lengths penalty of $0.3$ to generate the answers.
Each answer is composed of at most $20$ tokens.
The inference time of our model takes about $2$s to answer one question.

\section{Additional Ablations}
\paragraph{GNN Types.}
We experimented with different types of GNNs within our full model.
As depicted in \autoref{tab:ablation_gnn}, the combination of \modelname\, with APPNP \cite{appnp} led to the best overall performance compared to other GNNs such as GAT \cite{gat}, GCN \cite{gcn}, and SAGE \cite{sage}.

\paragraph{Mode Size.}
Moreover, we experimented with different sizes our model. 
As depicted in \autoref{tab:ablation_size}, the variant of \modelname\, that is based on BART-base significantly under-performed the large variant across all evaluation metrics of both datasets. 

\begin{table}[tb]
  \caption{Additional ablations of \modelname.}
  \centering
  \begin{subtable}{0.47\linewidth}
      \caption{Performance comparison of our best model using different GNN types.}
    \label{tab:ablation_gnn}    
        \centering
        \scalebox{0.78}[0.78]{
            
            \begin{tabular}{lcccccc}
            \toprule
              \multirow{2}*{\modelname} & \multicolumn{3}{c}{\textbf{AVSD-DSTC7}} & \multicolumn{3}{c}{\textbf{AVSD-DSTC8}} \\
              \cmidrule(r){2-4}   \cmidrule(r){5-7}
    
            & \textbf{B-4} & \textbf{R} & \textbf{C} & \textbf{B-4} & \textbf{R} & \textbf{C} \\
            \midrule
            w/ GAT   & $\underline{46.7}$  & $61.5$  & $135.4$  & $46.5$ & $60.9$ & $129.4$ \\
            w/ GCN   & $46.6$  & $\underline{61.9}$  & $\underline{136.7}$  & $\underline{46.7}$ & $\underline{61.6}$ & $\underline{131.6}$\\
            w/ SAGE  & $46.0$  & $61.2$  & $133.4$  & $45.8$ & $60.9$ & $129.3$\\
            w/ APPNP & $\mathbf{47.6}$ & $\mathbf{62.5}$  & $\mathbf{138.8}$  & $\mathbf{47.7}$  & $\mathbf{62.3}$ & $\mathbf{134.9}$ \\
            \bottomrule
            \end{tabular}
    }
   
  \end{subtable}
  \hfill
  \begin{subtable}{0.47\linewidth}
        \centering
        \caption{Performance comparison between different model sizes. ``Base'' and ``Large'' mean that \modelname\, uses a base or a large backbone, respectively.}
        \label{tab:ablation_size}
        \scalebox{0.78}[0.78]{
         
        \begin{tabular}{lcccccc}
        \toprule
          \multirow{2}*{\modelname} & \multicolumn{3}{c}{\textbf{AVSD-DSTC7}} & \multicolumn{3}{c}{\textbf{AVSD-DSTC8}} \\
          \cmidrule(r){2-4}   \cmidrule(r){5-7}
    
        & \textbf{B-4} & \textbf{R} & \textbf{C} & \textbf{B-4} & \textbf{R} & \textbf{C} \\
        \midrule
        Base ($\Delta=2$)   & ${39.8}$  & $60.0$  & $113.9$  & $40.1$ & $55.4$ & $110.2$ \\
        Large ($\Delta=4$)   & $\mathbf{47.6}$  & $\mathbf{62.5}$  & $\mathbf{138.8}$  & $\mathbf{46.7}$ & $\mathbf{61.6}$ & $\mathbf{131.6}$\\
        \bottomrule
        \end{tabular}
    }
  \end{subtable}
\end{table}

\section{Qualitative Results}
\label{appendix:qualitative_res}
\begin{figure}[h]
    \begin{subfigure}[b]{1\textwidth}
        \centering
        \scalebox{0.9}[0.9]{
            \includegraphics[width=\textwidth]{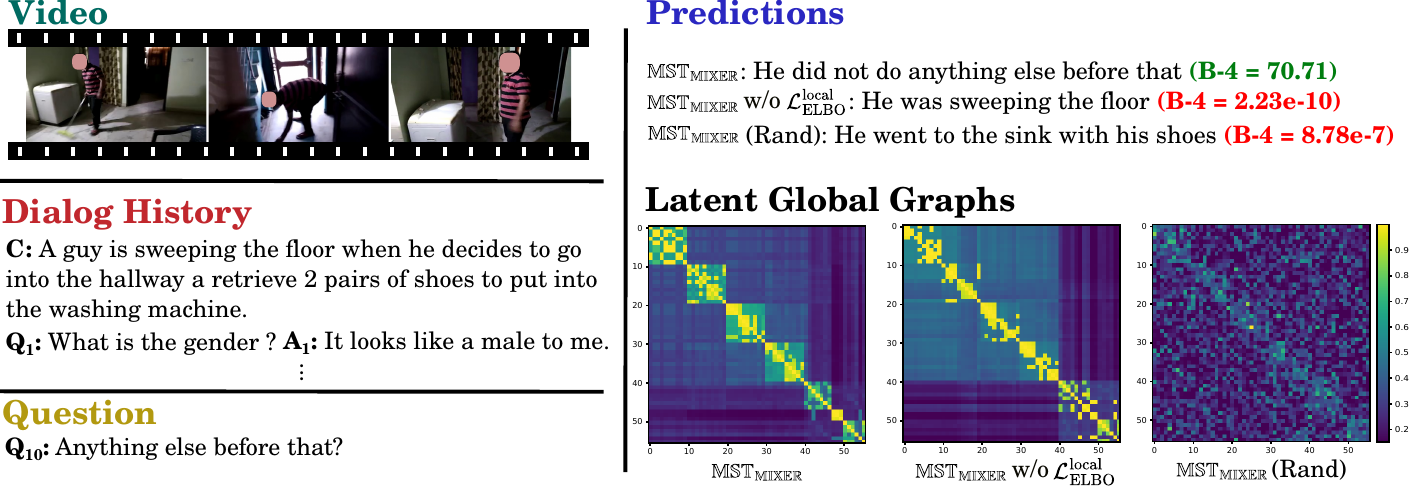}
        } 
        \caption{Dialog with video-id = 3A9IC.}
    \end{subfigure}
    \par\medskip
    
    \begin{subfigure}[b]{1\textwidth}
        \centering
        \scalebox{0.9}[0.9]{
            \includegraphics[width=\textwidth]{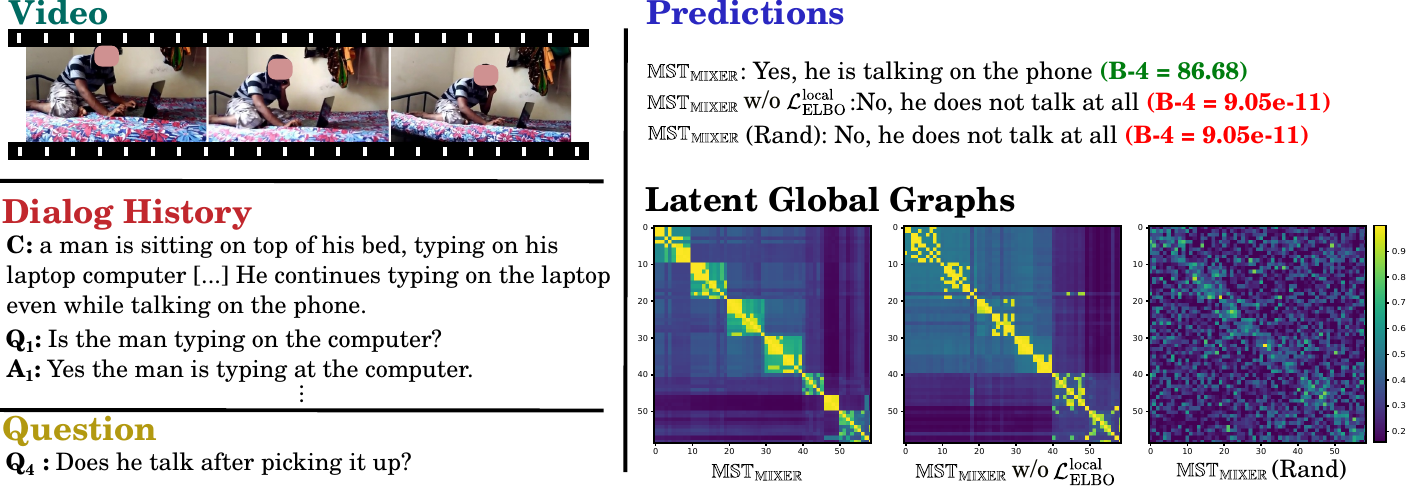}
        } 
        \caption{Dialog with video-id = 2EW71.}
    \end{subfigure}
    \par\medskip
    
        \begin{subfigure}[b]{1\textwidth}
        \centering
        \scalebox{0.9}[0.9]{
            \includegraphics[width=\textwidth]{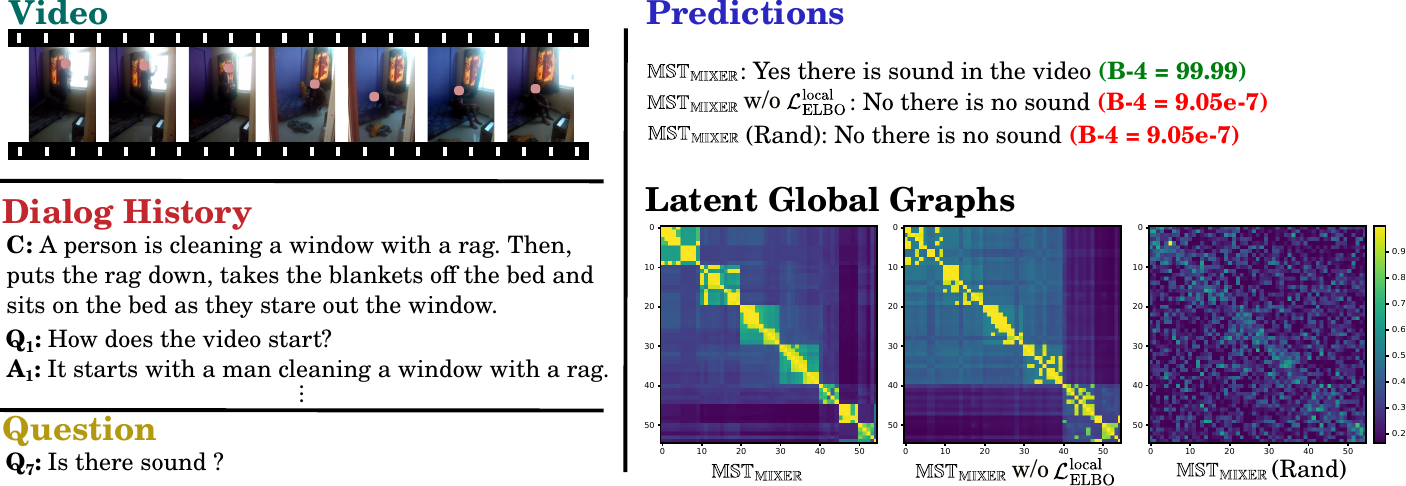}
        } 
        \caption{Dialog with video-id = K2XKT.}
    \end{subfigure}
    \par\medskip
    
    \begin{subfigure}[b]{1\textwidth}
        \centering
        \scalebox{0.9}[0.9]{
            \includegraphics[width=\textwidth]{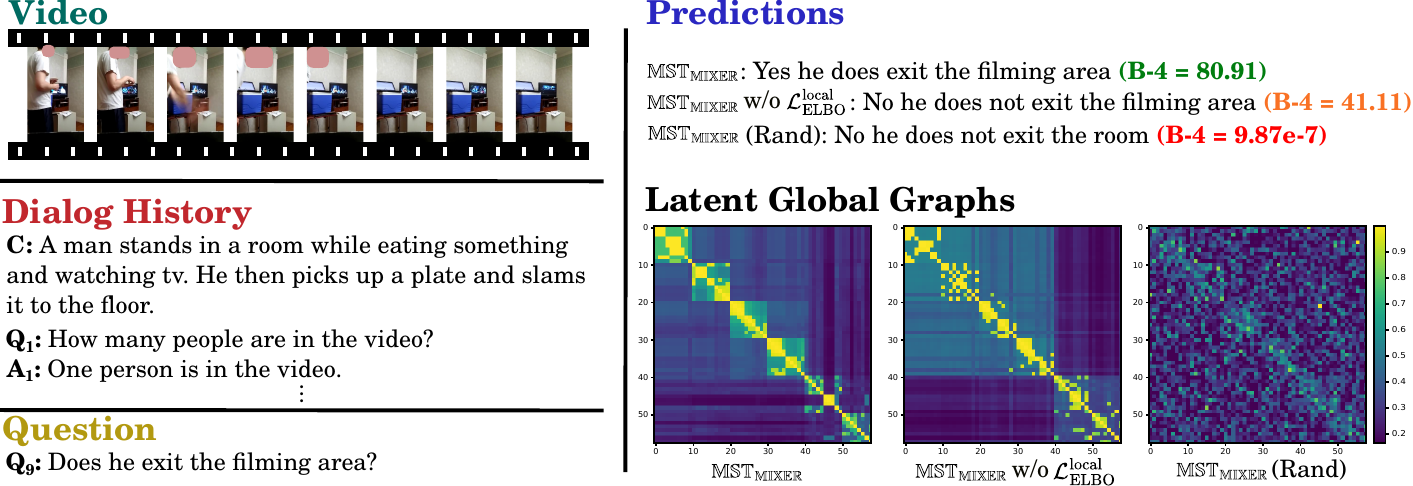}
        } 
        \caption{Dialog with video-id = UO7PC. Although the ablated version (\modelname w/o $\mathbf{\mathcal{L}_\textrm{ELBO}^\mathrm{local}}$) reached a BLEU-4 score of $41.11$, it incorrectly answered the question since the person did leave the filming area as can be seen from the last frames of the video.}
    \end{subfigure}

    \caption{Qualitative results on data samples form the test split of AVSD-DSTC7. For ethical reasons, we blurred the faces of people appearing in the video frames.}
    \label{fig:qualitative_2}
  \end{figure}

\begin{figure}[h]
    \begin{subfigure}[b]{1\textwidth}
        \centering

        \scalebox{0.9}[0.9]{
            \includegraphics[width=\textwidth]{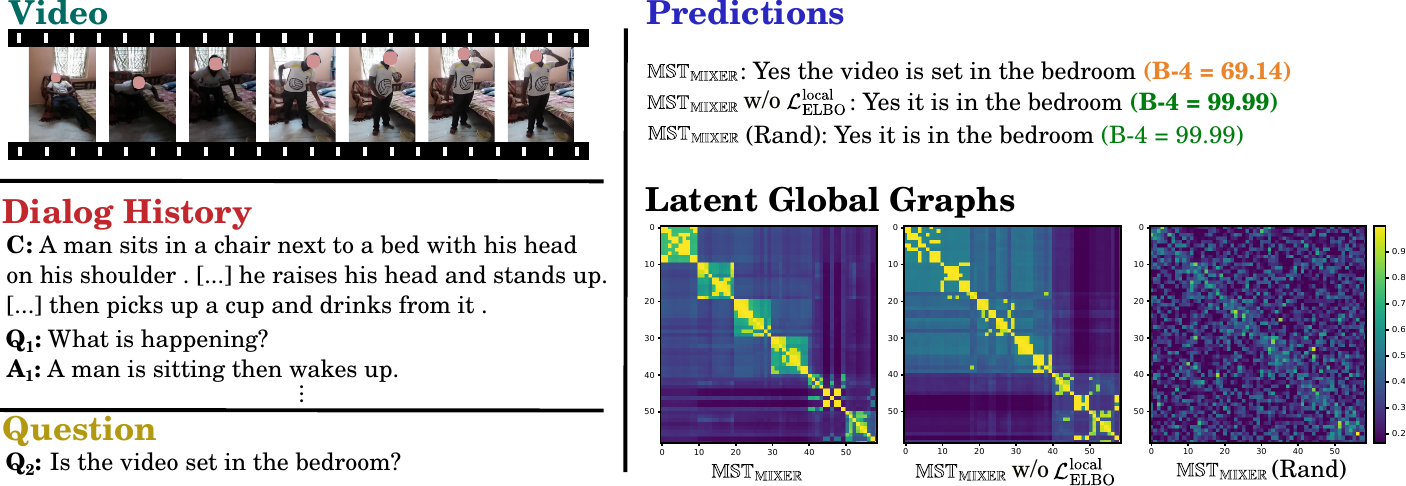}
        } 
        \caption{Dialog with video-id = 3DR7T.}
    \end{subfigure}
    
    \par\medskip
    
    \begin{subfigure}[b]{1\textwidth}
        \centering

        \scalebox{0.9}[0.9]{
            \includegraphics[width=\textwidth]{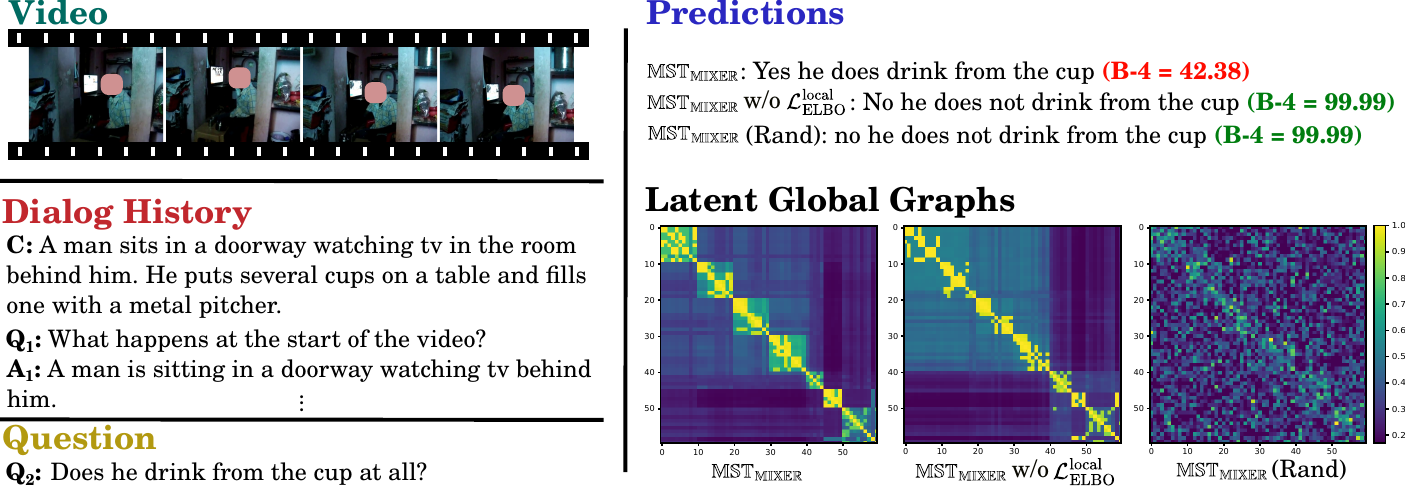}
        } 
        \caption{Dialog with video-id = HKGAX.}
    \end{subfigure}
    
    \par\medskip
    
    \begin{subfigure}[b]{1\textwidth}
        \centering

        \scalebox{0.9}[0.9]{
            \includegraphics[width=\textwidth]{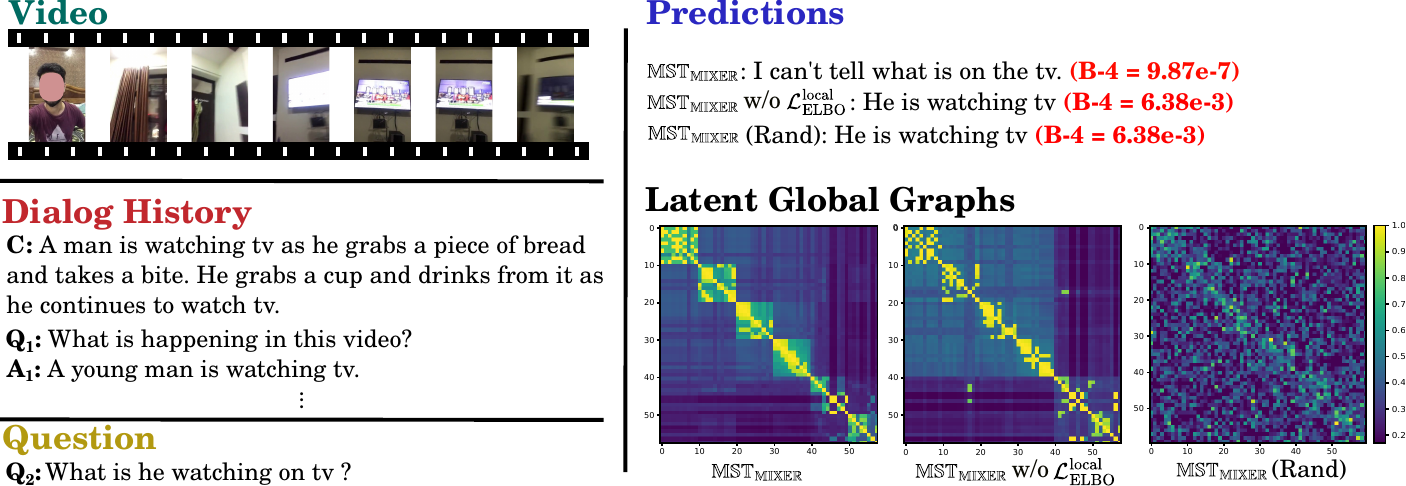}
        } 
        \caption{Dialog with video-id = 1K4NH.}
    \end{subfigure}

    \par\medskip
    
    \begin{subfigure}[b]{1\textwidth}
        \centering

        \scalebox{0.9}[0.9]{
            \includegraphics[width=\textwidth]{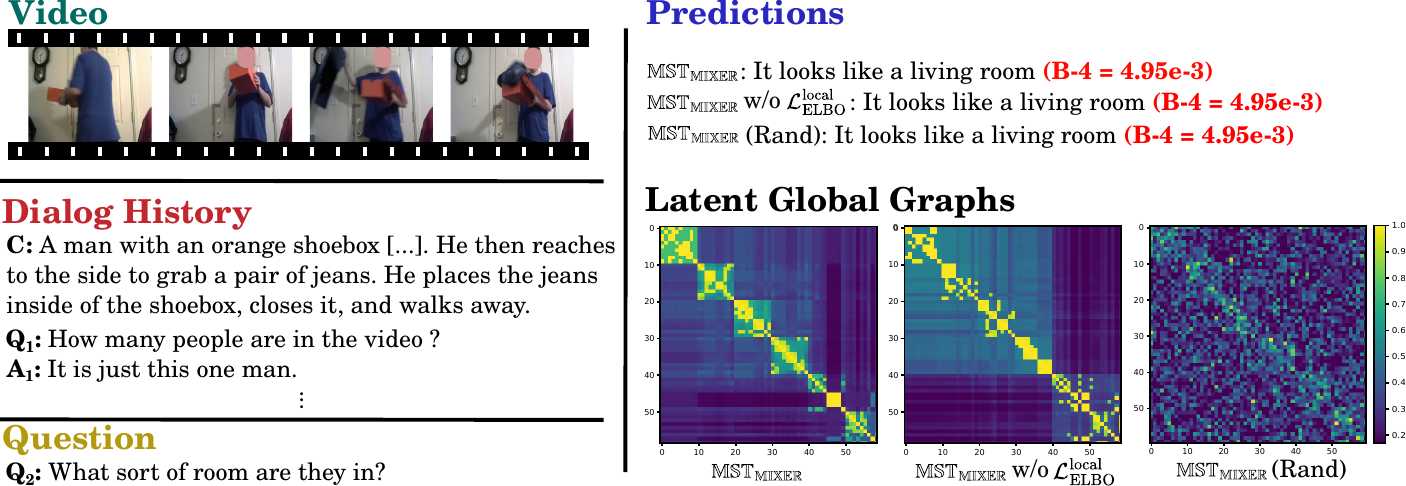}
        } 
        \caption{Dialog with video-id = 36QP8.}
    \end{subfigure}
    \caption{Negative qualitative results on data samples form the test split of AVSD-DSTC7. For ethical reasons, we blurred the faces of people appearing in the video frames.}
    \label{fig:qualitative_3}

  \end{figure}

We provide additional extensive qualitative examples of our best model and some of its ablated versions for comparison in \autoref{fig:qualitative_2}.
Finally, we give some failure cases in \autoref{fig:qualitative_3}.

\begin{figure}[!ht]
\begin{lstlisting}[language=Python, caption={PyTorch implementation of the ELBO loss. Since $q_\phi$ and $p_\theta$ only output normalized scores as the prediction for each edge, we append the zero vectors to both predictions in lines 29-30 to convert the raw scores to a two-value probability before applying softmax.}, label=lst:elbo]
# ---------------------------------
#  Implementation of the ELBO loss
# ---------------------------------
import torch
import torch.nn as nn
import torch.nn.functional as F

class ELBO(nn.Module):
    def __init__(self):
        super(ELBO, self).__init__()

    def forward(self, Aq, Ap):
        """
        Args:
            Aq: The predicted latent graph of q_phi 
                shape = (batch_size, K, K)   -- local graphs
                shape = (batch_size, NK, NK) -- global graphs

            Ap: The predicted latent graph of p_theta
                shape = (batch_size, K, K)   -- local graphs
                shape = (batch_size, NK, NK) -- global graphs
        
        Returns:
            The ELBO loss
        """
        Aq_flat = Aq.view(-1).unsqueeze(-1)
        Ap_flat = Ap.view(-1).unsqueeze(-1)

        Aq_flat = torch.cat(
            [torch.zeros_like(Aq_flat), Aq_flat], dim=-1)
        Ap_flat = torch.cat(
            [torch.zeros_like(Ap_flat), Ap_flat], dim=-1)
        
        log_Aq = F.log_softmax(QA_flattened, dim=1)
        log_Ap = F.log_softmax(PA_flattened, dim=1)

        Aq_dist = torch.exp(log_Aq)

        loss_Aq = torch.mean(log_Aq * Aq_dist)
        loss_Ap = torch.mean(log_Ap * Aq_dist)

        elbo_loss = loss_Aq - loss_Ap

        return elbo_loss
        \end{lstlisting}
\end{figure}

\begin{table}[!t]
    \centering
    \caption{Detailed hyperparameter setting of the training and inference of our best \modelname\, model.}
    \scalebox{0.83}[0.83]{
        \begin{tabular}{llc}
        \toprule
        \multirow{1}*{\textbf{Category}} & \multirow{1}*{\textbf{Hyperparameter}}  &  \\
        \midrule
        \multirow{19}*{\textbf{Model Architecture}}
                            & Dimension of I3D rgb / I3D flow / SAM features $d_v$ & $2048$   \\
                            & Dimension of SAM features $d_s$ & $512$   \\
                            & Maximum length of I3D rgb / I3D flow / SAM features $d_l$ & $36$ \\
                            & Dimension of audio features $d_a$ & $128$   \\
                            & Maximum length of audio features $l_a = l_v$ & $36$   \\
                            & Maximum total length of multi-modal input & $1024$ \\
                            & Dimension of hidden features $d$ & $1024$/$768$ \\
                            & Number of node features in local GNNs $K$ & $10$ \\
                            & Number of node features in local GNNs $K$ & $10$ \\
                            & Number for kNNs in $\{\tilde{A}_i\}$ & $4$ \\
                            & Number of learnable weights of \textcolor{red}{Equation 7} & $8$ \\
                            & Input dimension of GNNs in \autoref{tab:ablation_gnn} & $1024$ \\
                            & Output dimension of GNNs in \autoref{tab:ablation_gnn} & $1024$ \\
                            & $K$ value of APPNP & $2$ \\
                            & $\alpha$ value of APPNP & $0.1$ \\
                            & Number of attention heads in local GATs & $2$ \\
                            & Number of attention heads in global GATs & $4$ \\
                            & $\lambda$ value & $0.9$ \\
                            & $\Delta$ value & $4$ \\

        \midrule
        \multirow{9}*{\textbf{Optimization}}
                            & Optimizer & AdamW \\
                            & Learning rate of parameters in the VLM backbone $\mathrm{lr}_{\mathrm{BART}}$ & 1e-5 \\
                            & Learning rate of other parameters $\mathrm{lr}_{\mathrm{rest}}$ & 1e-4 \\
                            & Values of $\{\alpha_1, \alpha_2, \alpha_3\} $  & $\{1, 100, 100\}$ \\
                            & Learning rate schedule & linear \\
                            & Dropout rate & $0.1$ \\
                            & Value of gradient clipping & $1.0$ \\
                            & Effective batch size & $96$ \\
                            & Number of epochs & $12$ \\
        \midrule
        \multirow{3}*{\textbf{Hardware}}
                            & GPU model & \texttt{Tesla V100-32GB} \\
                            & Number of GPUs  & $8$ \\
                            & Distributed training &\texttt{PyTorch DDP} \\
        \midrule
        \multirow{4}*{\textbf{Inference}}
                            & Maximum number of response tokens & 20 \\
                            & Depth of beam search  & $5$ \\
                            & Length penalty in beam search & $0.3$ \\
                            & Batch size & $1$ \\

        \bottomrule
        
        \end{tabular}
    }
    \label{tab:hyperparams_1}
\end{table}


%
%
\bibliographystyle{splncs04}
\bibliography{main}
\end{document}